\documentclass[10pt,twocolumn,letterpaper]{article}

\usepackage[pagenumbers]{cvpr} 

\usepackage[utf8]{inputenc} 
\usepackage[T1]{fontenc} 
\usepackage{microtype} 
\usepackage{amsmath}

\usepackage{graphicx} 
\usepackage{subcaption} 
\usepackage{wrapfig} 
\usepackage{placeins}
\usepackage{stfloats}
\usepackage{booktabs} 
\usepackage{array}
\usepackage{longtable}
\usepackage{multirow} 
\usepackage{makecell} 
\usepackage{xcolor} %
\usepackage{colortbl} %

\definecolor{OpenModel}{RGB}{240,255,240}
\definecolor{ClosedModel}{RGB}{248,245,252}
\newcommand{\oc}{\cellcolor{OpenModel}}
\newcommand{\cc}{\cellcolor{ClosedModel}}
\definecolor{TextOnly}{RGB}{240,255,240}
\definecolor{ImageText}{RGB}{245,242,249}

\definecolor{AudioText}{RGB}{237,244,250}
\definecolor{TriModal}{RGB}{252,245,235}

\definecolor{humanreview}{RGB}{237,249,247}
\definecolor{llmjudge}{RGB}{252,243,247}
\usepackage{enumitem} 
\usepackage{pifont} 
\usepackage{fontawesome} 
\usepackage{titletoc} 
\usepackage{caption}

\usepackage[table]{xcolor}

\usepackage[most]{tcolorbox}

\newtcolorbox{chatbox}{
  enhanced,
  breakable,
  colback=white,
  colframe=black!70,
  boxrule=0.8pt,
  arc=4pt,          
  left=10pt,right=10pt,top=10pt,bottom=10pt,
}

\newcommand{\methodname}{BenchHarness} 
\definecolor{cvprblue}{rgb}{0.21,0.49,0.74}
\usepackage[pagebackref,breaklinks,colorlinks,allcolors=cvprblue]{hyperref}

\def\confName{CVPR}
\def\confYear{2027}

\title{Benchmark Everything Everywhere All at Once}

\newcommand{\aff}[1]{\textsuperscript{#1}}
\author{%
    \begingroup
    \renewcommand{\arraystretch}{1.02}
    \begin{tabular}{c}
    \textbf{Shiyun Xiong\aff{1}\thanks{Equal contribution.} \quad
    Dongming Wu\aff{1}\aff{,2}\footnotemark[1] \quad
    Peiwen Sun\aff{1} \quad
    Yuang Ai\aff{1}} \\
    \textbf{Bokang Yang\aff{3}\aff{,4} \quad
    Wencheng Han\aff{5} \quad
    Xiao-Hui Li\aff{6} \quad
    Xiangyu Yue\aff{1}\aff{,4}\thanks{Corresponding author.}} \\[0.5em]
    \normalfont \aff{1}MMLab, The Chinese University of Hong Kong \quad
    \normalfont \aff{2}CPII under InnoHK \\
    \normalfont \aff{3}The Chinese University of Hong Kong, Shenzhen \quad
    \normalfont \aff{4}Shenzhen Loop Area Institute \\
    \normalfont \aff{5}Shandong University \quad
    \normalfont \aff{6}Huawei Technologies \\
    \end{tabular}
    \endgroup
}

\makeatletter
\let\originalmaketitle\@maketitle
\renewcommand{\@maketitle}{%
    \originalmaketitle
    \par\noindent
    \begin{minipage}{\textwidth}
        \centering
        \includegraphics[width=\linewidth]{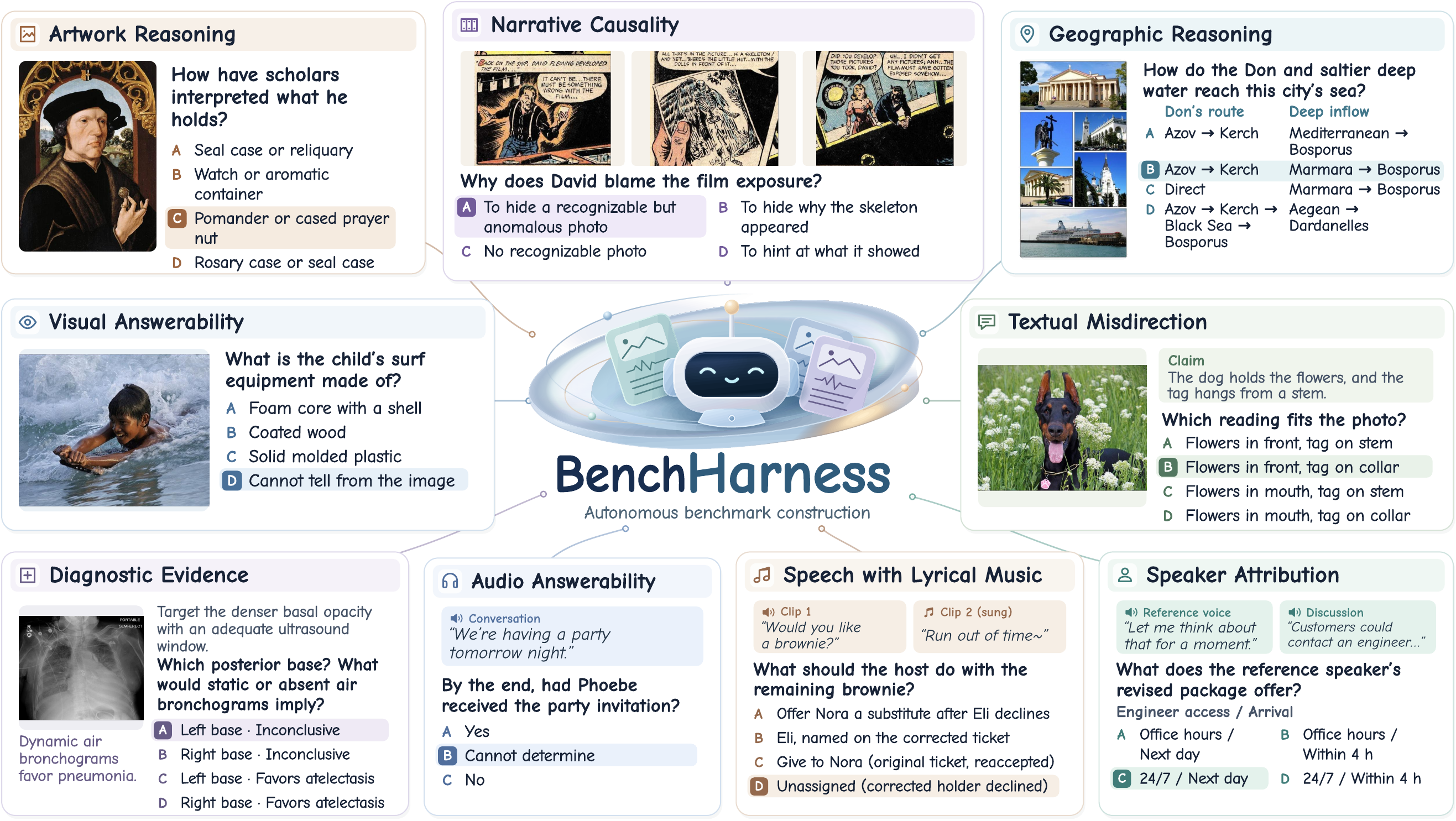}
        \vspace{-4pt}
        \captionof{figure}{\textbf{Benchmark customization with \methodname{}.}
Highlighted choices indicate reference answers, and audio inputs are shown as text excerpts. BenchHarness is an end-to-end harness that autonomously builds high-quality benchmarks tailored to user-specified evaluation goals. The examples span visual and audio tasks, from artwork interpretation to speaker attribution.}
        \label{fig:main_image}
    \end{minipage}
    \par\medskip
}
\makeatother
\begin{document}

\maketitle

\begin{abstract}
Benchmarks provide standardized measures for evaluating and advancing large language models (LLMs) and multimodal LLMs (MLLMs). As models improve, existing benchmarks can approach saturation, making it harder to challenge models and distinguish among them. While reusing annotated datasets can reduce data collection and labeling effort, constructing new benchmarks remains labor-intensive, and existing agent-based approaches often target specific tasks or automate only parts of the process.
To address these challenges, we introduce \textbf{\methodname{}}, an autonomous harness for end-to-end benchmark customization. It coordinates multiple agents and tools to analyze user requirements, search for suitable datasets, and construct validated benchmarks, with support for refinement based on model performance. 
To evaluate \methodname{}, we construct \textbf{16 benchmarks} spanning diverse tasks and modalities. Human review and LLM-as-a-judge assessments indicate high benchmark quality, while evaluations across multiple models reveal challenging tasks and meaningful performance differences. In comparisons with source benchmarks, the generated benchmarks yield lower scores and larger performance gaps, even when the originals approach saturation. \methodname{} thus provides a reusable approach to benchmark construction, supporting informative evaluation as model capabilities and user needs evolve. Our code and benchmarks are publicly available at \href{https://github.com/Shiyun-x/BenchHarness}{\faGithub \ Github}.


\end{abstract}    
\section{Introduction}
With the rapid advancement of large language models (LLMs), such as ChatGPT~\cite{openai2026gpt6astra}, Gemini~\cite{google2026gemini4carbon}, and Claude~\cite{anthropic2026fable52}, their capabilities have expanded from basic understanding to complex reasoning and from single-modal processing to multimodal comprehension. Throughout this evolution, benchmarks have played a central role by providing standardized tasks, datasets, and evaluation protocols that track the progress of different approaches~\cite{berkeleyrdi2026whyagentslastexam}. As the range of real-world applications continues to expand across diverse domains and task settings, more comprehensive or fine-grained evaluation is needed.

\begin{figure}[t]
    \centering
    \includegraphics[width=\linewidth]{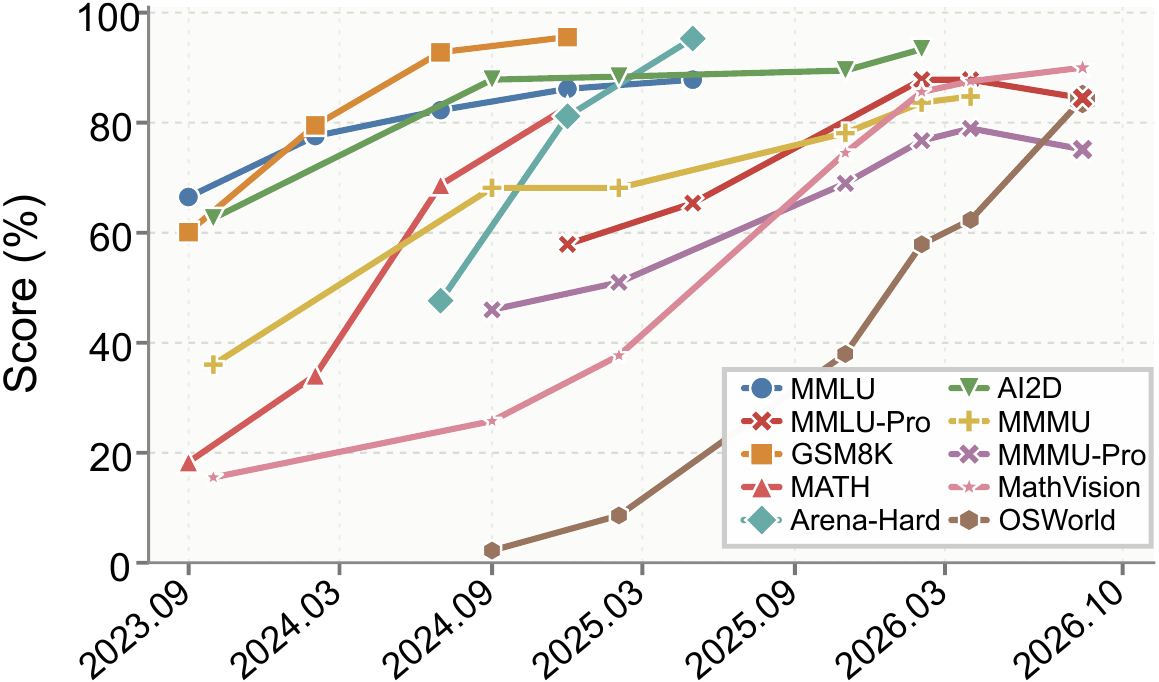}
    \vspace{-4pt}
    \caption{\textbf{Benchmark performance saturation on Qwen}.}
    \label{fig:motivation}
    \vspace{-10pt}
\end{figure}

However, as model capabilities continue to advance, static benchmarks can rapidly approach performance saturation~\cite{pmlr-v306-akhtar26a}. As illustrated in Figure~\ref{fig:motivation}, scores on several representative benchmarks rise markedly across successive Qwen and Qwen-VL releases~\cite{qwen,qwen2,qwen2vl,qwen25vl,qwen35blog,qwen36plus}, leaving less room to distinguish among high-performing models. This motivates a continuing need for new benchmarks that evaluate emerging capabilities and enable more informative model comparisons. Nevertheless, constructing new benchmarks often involves collecting, curating, and annotating data, which can be time-consuming and labor-intensive~\cite{wang2024mmlu,sun2025spacevista,yang2026mmsi,hong2025motionbench,fu2026video}.

One promising direction is to repurpose existing annotated datasets and benchmarks, drawing on their rich, vetted information for new evaluation goals without collecting data from scratch.
Prior work has used general datasets to construct benchmarks targeting specific capabilities~\cite{bhat2026bopask,Hudson_2019_CVPR,sakshi2025mmau,bhattacharya2025benchmarking}. Domain-specific data have also been used to develop benchmarks for various tasks in fields including medicine~\cite{NEURIPS2025_7ef1f85b,hu2023expert}, agriculture~\cite{Shinoda_2025_ICCV,quoc2026leafnet}, and finance~\cite{reddy-etal-2024-docfinqa,tang-etal-2025-financereasoning,strich2026t2}. Even with annotated data, constructing benchmarks often requires substantial human effort in task design and quality control~\cite{yang2026mmsi,hong2025motionbench}. Recent LLM- and agent-based approaches have been developed to automate benchmark construction, including pipelines for specific tasks or data types~\cite{dutta2026condabench,gwiazda2026timeseriesexamagent,shashidhar2025yourbench} and methods that evolve existing benchmark instances~\cite{wang-etal-2025-benchmark,liu2025arenabencher}. 
However, these approaches typically target specific tasks or automate only part of benchmark construction, leaving the full process of analyzing user requirements, searching for suitable data sources, and constructing validated benchmarks from them less explored.

To address these challenges, we introduce \textbf{\methodname}, an end-to-end harness that coordinates multiple agents and tools to automatically build benchmarks for user-specified evaluation needs.
Inspired by the brain--cerebellum hierarchical architecture~\cite{ren2025neuronal,zhao2025dual,israely2025cerebellar,tan2025roboos}, our harness adopts a dual-component design. 
The \textit{Benchmark Planner}, a high-level decision module, translates user requirements into concrete, feasible benchmark specifications. 
The \textit{Benchmark Executor}, an operational module, turns these specifications into standardized, runnable benchmarks. 
Furthermore, \methodname{} supports benchmark refinement based on models' performance.
By integrating planning, execution, and refinement, \methodname{} consistently aligns benchmark construction with high-level evaluation requirements while accounting for real-world data conditions and execution constraints.

\methodname~ has three key features. 
\ding{182} \textit{User-oriented Customization.}
While many existing benchmarks are built around predefined evaluation tasks~\cite{hendrycks2020measuring,wang2024mmlu,liu2024mmbench,yue2025mmmu} and most automated construction methods rely on preselected data sources~\cite{dutta2026condabench,gwiazda2026timeseriesexamagent}, \methodname{} identifies suitable data and tailors benchmarks to user-specified requirements across domains and modalities.
\ding{183} \textit{High-quality Benchmarks.} 
The constructed benchmarks provide reliable and challenging assessments of the intended capabilities, revealing model limitations and enabling informative comparisons.
\ding{184} \textit{Automated and Reusable Construction.} 
The harness automates the full construction process through a reusable workflow, reducing human effort and enabling generation as models improve and evaluation needs evolve.

To evaluate \methodname{}, we conduct extensive experiments across 16 constructed benchmarks spanning diverse tasks and modalities.
First, we assess benchmark quality and reliability through human evaluation and LLM-as-a-judge assessment.
Second, we evaluate multiple models on these benchmarks to examine their difficulty and discriminative ability. We further compare constructed benchmarks with their source benchmarks to assess whether the new tasks provide informative evaluations when the originals approach saturation.
Third, we compare benchmarks built by each backbone model with and without the harness to assess its effectiveness, and ablate each component to examine its contributions.
Overall, our key contributions are follows:
\begin{itemize}
    \item We present the \textbf{performance saturation} of existing benchmarks and highlight the need for timely and rapid benchmark updates to expose model limitations and distinguish model performance. 
    \item We propose \textbf{\methodname{}}, an autonomous harness that coordinates multiple agents and tools to analyze user-specified requirements, search for relevant existing data, and transform them into customized benchmarks.
    \item We construct \textbf{16 benchmarks} spanning diverse tasks and modalities. Extensive experiments demonstrate their quality and evaluation value, as well as the effectiveness of \methodname{} and its components.
\end{itemize}

\section{Related Works}
\noindent\textbf{Model Evaluation Benchmarks.}
Benchmarks provide standardized assessments of language and multimodal models. General-purpose benchmarks assess knowledge, perception, and reasoning across diverse subjects and tasks~\cite{hendrycks2020measuring,wang2024mmlu,liu2024mmbench,yue2025mmmu}. Others focus on specific capabilities or model behaviors, such as spatial and temporal reasoning and multimodal understanding under particular conditions~\cite{bhat2026bopask,yang2026mmsi,sakshi2025mmau,bhattacharya2025benchmarking}. For example, M3-SLU~\cite{kwon-etal-2026-m3} assesses multi-speaker conversation understanding, POPE~\cite{li-etal-2023-evaluating} probes object hallucination in vision-language models, and ConText-VQA~\cite{zhang-etal-2026-images} examines their susceptibility to misleading textual context. In specialized domains, benchmarks address tasks such as medical image question answering~\cite{NEURIPS2025_7ef1f85b,hu2023expert}, crop disease identification in agriculture~\cite{Shinoda_2025_ICCV,quoc2026leafnet}, and financial document reasoning~\cite{reddy-etal-2024-docfinqa,tang-etal-2025-financereasoning,strich2026t2}. Together, these benchmarks illustrate how evaluation goals vary across capabilities, conditions, and domains, highlighting the value of designing benchmarks for specific needs.

\noindent\textbf{Benchmark Construction with LLMs.}
Recent work has explored the use of LLMs in benchmark construction. 
In human-led efforts, LLMs assist with item generation, filtering, or revision, while researchers remain involved in curation and verification~\cite{yue2025mmmu,xuan-etal-2025-mmlu,hong2025motionbench,liu2024mibench,chen2024we,wang2024mmlu}.
Other approaches automate construction for specific tasks or types of source data~\cite{gwiazda2026timeseriesexamagent,malaguti-etal-2025-orqa,guo2025swe,fu2026medbenchagent}. For example, AutoPaperBench~\cite{electronics14061175} generates questions to evaluate scientific paper understanding, while ConDABench~\cite{dutta2026condabench} creates interactive data analysis tasks from articles and their associated datasets. 
More general frameworks further automate benchmark design, generation, or evolution across evaluation goals~\cite{butt2024benchagents,shashidhar2025yourbench,wang-etal-2025-benchmark,bouzenia2026benchmeup}.
Among them, AutoBencher~\cite{ICLR2025_eb216114} searches for evaluation topics and grounds knowledge-intensive questions in retrieved documents, whereas ArenaBencher~\cite{liu2025arenabencher} uses feedback from multiple models to evolve existing items while preserving their evaluation objectives. 
Despite these advances in automating benchmark construction, end-to-end approaches that identify suitable datasets and adapt them to diverse user-specified evaluation goals remain less explored.

\noindent\textbf{Agentic Harness for Long-Horizon Work.}
Early LLM agents combined reasoning, tool use, and environmental feedback to perform tasks iteratively~\cite{yao2022react,shinn2023reflexion,wang2023voyager}.
As tasks grew more complex, subsequent frameworks developed agentic workflows that coordinate model calls and tools through task decomposition and multi-agent collaboration~\cite{wu2023autogen,hong2024metagpt,gao2024agentscope,pmlr-v235-kim24y}.
For example, Magentic-One~\cite{fourney2024magentic} uses an orchestrator to plan tasks, track progress, and direct specialized agents, while Agent Laboratory~\cite{schmidgall-etal-2025-agent} coordinates literature review, experiments, and report writing.
LongHorizon-Harness~\cite{ma2026longhorizon} further maintains task state and audits intermediate results to support long-horizon execution.
Agentic workflows have also been used to construct data.
AgentInstruct~\cite{mitra2024agentinstruct} generates instruction data from text and code, while APIGen-MT~\cite{prabhakar2026apigen} synthesizes tool-use trajectories through simulated interactions.
Omni-Detective~\cite{ma2026omni} also uses tools to generate fine-grained descriptions of audio and video.
Benchmark construction similarly involves interdependent stages, from task design and dataset grounding to item generation and quality control. Coordinating these stages within an agentic harness to construct reliable benchmarks at scale remains a challenge.

\section{\methodname}
\begin{figure*}
    \centering
    \includegraphics[width=\linewidth]{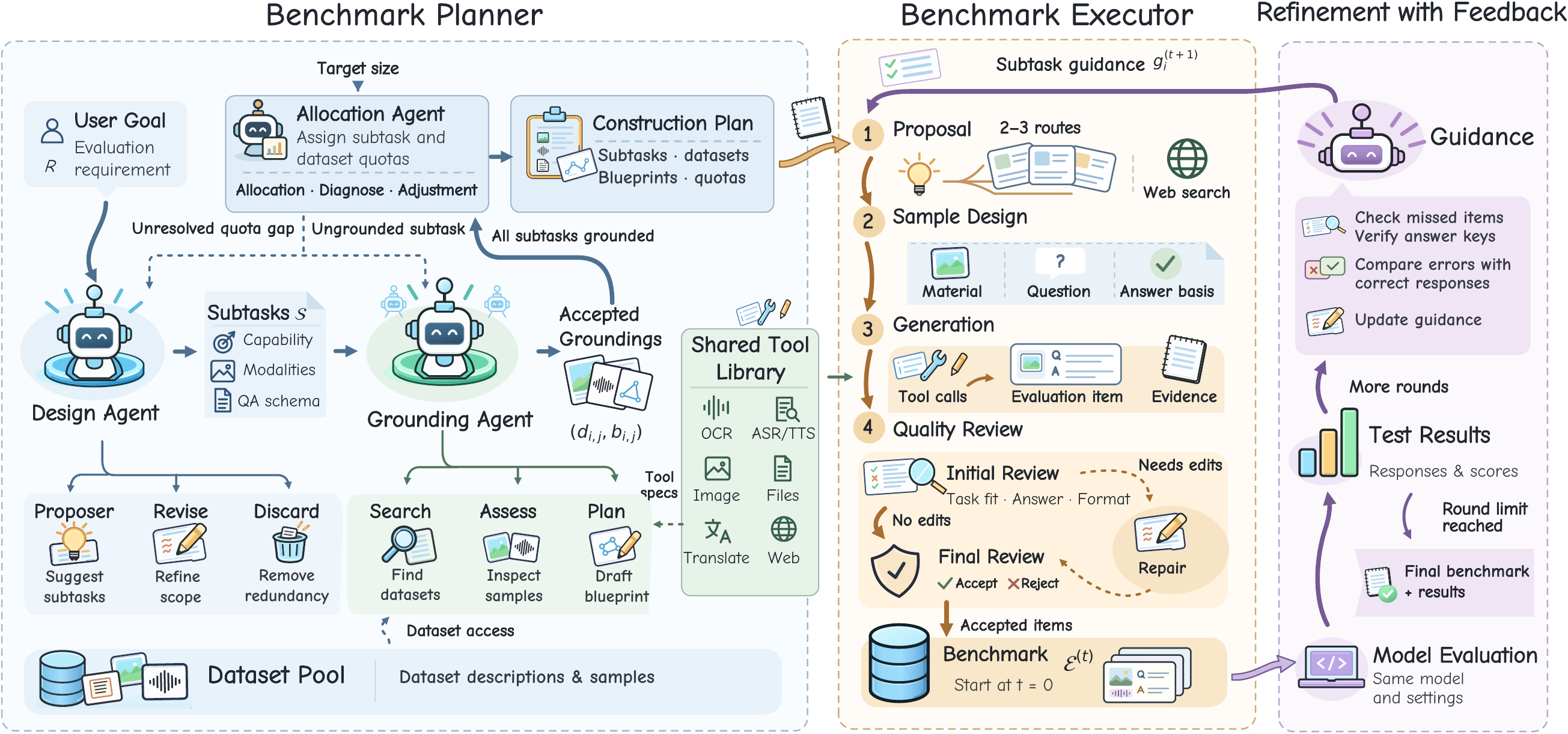}
    \caption{The overall pipeline of \methodname{}. The \textbf{Benchmark Planner} defines subtasks, investigates and selects datasets, produces dataset-level blueprints, and assigns sample quotas.
The \textbf{Benchmark Executor} develops sample-level designs, constructs items using available tools, and reviews and refines them for inclusion in the benchmark.}
    \label{fig:method}
    \vspace{-8pt}
\end{figure*}
In this section, we present \methodname, an end-to-end harness that coordinates multiple agents and tools to construct benchmarks for user evaluation requirements, as illustrated in Figure~\ref{fig:method}.
Given an abstract evaluation goal and a target size, the \textbf{Benchmark Planner} defines subtasks, selects suitable datasets, outlines their potential uses in blueprints, and assigns sample quotas. The \textbf{Benchmark Executor} uses the resulting plan to design and construct items with available tools, then reviews them to form the initial benchmark. When the benchmark is used to evaluate a model, the system can analyze its responses to guide another round of item construction.
We detail each component below; prompt templates are provided in Appendix~\ref{supp:prompt_templates}.
 
\subsection{Benchmark Planner}
User requirements, \textit{e.g.}, \textit{"I want to evaluate whether a model can understand a target speaker amid overlapping speech."}, are typically user-oriented.
The Benchmark Planner translates a requirement $R$ into a set of testable subtasks $\mathcal{S}=\{s_i\}_{i=1}^{N}$ and identifies at least one feasible grounding for each subtask. It consists of three collaborating agents: the Design Agent, the Grounding Agent, and the Allocation Agent. For each subtask $s_i$, let $\mathcal{D}_i=\{d_{i,j}\}_{j=1}^{M_i}$ denote its candidate dataset set.
A grounding for $s_i$ is represented as $(d_{i,j}, b_{i,j})$, where $d_{i,j}$ is a selected dataset and $b_{i,j}$ is a dataset-level blueprint outlining how its data can support the evaluation objective of $s_i$. 

\textbf{Design Agent.}
The Design Agent develops subtasks $\mathcal{S}$ from $R$, selecting tools as needed. 
It uses \texttt{Proposer} to generate candidate subtasks and decides which to retain based on what each one tests and how it contributes to the evaluation goal. The agent may invoke \texttt{Revise} to adjust a subtask's formulation or scope, \texttt{Discard} to remove a redundant subtask or one with unclear evaluation value, or \texttt{Proposer} again to address gaps in coverage. Once the subtasks are clearly defined and collectively cover the requested capabilities, the agent passes $\mathcal{S}$ to the Grounding Agent.

\textbf{Grounding Agent.}
For each subtask $s_i$, the Grounding Agent identifies datasets whose contents could support the evaluation objective using the available transformation tools and summarizes their potential in dataset-level blueprints. It selects and invokes the tools below as needed.

\textit{i) Dataset Search.}
When the workspace has no candidate datasets for $s_i$ or the agent needs more, it invokes \texttt{Search} with guidance on what to look for. \texttt{Search} first calls \texttt{Preference}, which combines this guidance with the task requirements to describe suitable data in terms of modality, annotation structure, and domain. Using these descriptions, \texttt{Search} retrieves candidates for $\mathcal{D}_i$, including datasets designed for similar tasks and other sources whose material could be adapted to $s_i$.

\textit{ii) Material Assessment.}
With candidate datasets in the workspace, the agent reviews their descriptions and may invoke \texttt{Assess} whenever it has specific questions about a candidate $d_{i,j}$. The tool consults dataset information and examines representative samples to answer these questions, helping the agent assess the dataset's potential for $s_i$.

\textit{iii) Blueprint Development.}
If the agent considers a candidate dataset $d_{i,j}$ promising for $s_i$, it may invoke \texttt{Plan} to develop a blueprint, providing guidance on what to explore. The tool draws on this guidance, the source material, any assessment findings, and descriptions of the Executor's transformation tools (Appendix~\ref{supp:transformation_tools}) to draft $b_{i,j}$. The blueprint outlines potential uses of the data and relevant tools, leaving the design of individual items to the Executor.

The agent reviews the blueprints and may accepts suitable $(d_{i,j},b_{i,j})$ pairs as groundings.  Once every subtask has at least one accepted grounding, the selected dataset--blueprint pairs are passed to the Allocation Agent. If a subtask cannot be grounded, $\mathcal{S}$ is returned to the Design Agent for revision.

\textbf{Allocation Agent.}
Given the accepted groundings and the target size, the Allocation Agent uses \texttt{Allocation} to assign sample quotas across subtasks and datasets, considering both subtask importance and dataset capacity. If an allocation leaves quotas unmet, the agent can use \texttt{Diagnose} to identify bottlenecks and use \texttt{Adjustment} to revise subtask priorities, dataset ordering, or quotas before reallocating. If these adjustments cannot resolve the shortfall, the system uses the feedback to revisit design and grounding.
The resulting construction plan is:
\[
\mathcal{B}={(s_i,\mathcal{G}_i)}_{i=1}^{N},
\]
where $\mathcal{G}_i$ contains the allocated tuples $(d_{i,j},b_{i,j},q_{i,j})$ for $s_i$, and $q_{i,j}$ denotes the number of source samples initially allocated from $d_{i,j}$. The Planner passes $\mathcal{B}$ to the Executor for sample-level design, construction, and review.

\subsection{Benchmark Executor}
\label{sec:benchmark_realization}
The Benchmark Executor uses the construction plan $\mathcal{B}$ to produce evaluation items for each subtask $s_i$.

\textbf{Benchmark Construction.}
For each $(d_{i,j},b_{i,j},q_{i,j})\in\mathcal{G}_i$, the Executor starts with the $q_{i,j}$ source samples allocated from $d_{i,j}$. The blueprint $b_{i,j}$ provides an initial direction that the Executor can adapt or replace to suit each sample.

\textit{i) Proposal.}
For each sample, the Executor explores two or three ways to develop its contents into an item for $s_i$. It may use targeted web searches to develop promising ideas or find support for additional factual claims.

\textit{ii) Design.}
Drawing on the proposals and supporting evidence, the Executor develops a sample-level design specifying the task material, question, and basis for the answer. It also describes how solving the item requires the target capability. For multiple-choice items, the design identifies what distinguishes the correct answer from plausible alternatives.

\textit{iii) Generation.}
Following the sample-level design, the Executor uses available transformation tools as needed to construct the complete item, including the material, question, reference answer, and answer options where applicable. It retains supporting evidence for review.

\textit{iv) Quality Review.}
An initial review checks whether the item aligns with $s_i$, has a valid answer, and follows the required format. If changes are needed, \texttt{Repair} revises the draft once using the review feedback. A final inspection then determines whether to accept the item.

If the number of accepted items falls short of the target size, the Executor can process unused source samples in a limited number of additional rounds and reports any remaining shortfall. The accepted items form the initial benchmark.

\textbf{Refinement with Feedback.}
Once the initial benchmark $\mathcal{E}^{(0)}$ is used to evaluate a model, its responses can guide further item construction.

\textit{i) Response Analysis.}
In round $t$, the Guidance Agent examines the model's responses to $\mathcal{E}^{(t-1)}$ for each subtask $s_i$. It checks incorrectly answered items and their reference answers for validity, then compares incorrect responses on valid items with correct responses to similar items. This comparison helps identify properties of the material or relationships among clues that may explain the errors.

\textit{ii) Guidance Development.}
The Guidance Agent summarizes the findings as guidance $g_i^{(t)}$ for $s_i$, highlighting cues or distinctions worth exploring in new items. The guidance sets a direction while leaving the Executor to decide how to adapt each source sample.

\textit{iii) Construction and Retesting.}
The Executor reruns the item construction process on the original source pool, using $g_i^{(t)}$ to guide sample-level design. After quality review, the accepted items form $\mathcal{E}^{(t)}$, on which the same model is retested under the original settings. In the next round, the Guidance Agent repeats the analysis using the new responses, taking $g_i^{(t)}$ into account when developing new guidance.

The process returns the resulting benchmark after up to the specified number of refinement rounds.
Through planning, construction, and refinement based on model responses, \methodname{} produces an evaluation-ready benchmark tailored to the user's requirements.

\newlength{\maintextfloatsep}
\setlength{\maintextfloatsep}{\textfloatsep}
\raggedbottom
\section{Experiments}
\label{sec:experiments}

\setlength{\textfloatsep}{10pt plus 2pt minus 2pt}

\definecolor{BenchmarkTitleRow}{HTML}{E8F1FA}
\definecolor{ModelComparisonGroupRow}{HTML}{EEF2F5}
\definecolor{ResultRowShade}{HTML}{F6F8FA}

\makeatletter
\newcommand{\benchmarkrowbackground}[1]{%
  \noalign{%
    \begingroup
      \color{#1}%
      \dimen@=\dimexpr\ht\@arstrutbox+\dp\@arstrutbox\relax
      \hrule height\dimen@
      \kern-\dimen@
    \endgroup
  }%
}
\makeatother

\begin{figure}[!b]
    \centering
    \includegraphics[width=\linewidth]{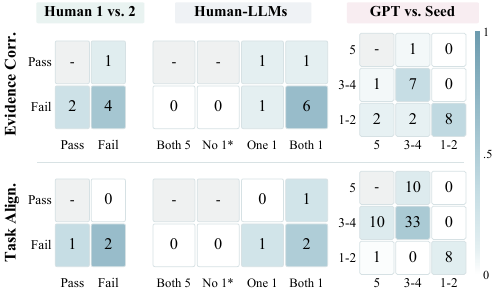}
    \vspace{-16pt}
    \caption{\textbf{Agreement between human reviewers and LLM judges.} Numbers show item counts, and colors indicate proportions after excluding gray cells.}
    \label{fig:judge_agreement}
\end{figure}

\begin{table*}[t]
\centering
\caption{\textbf{\methodname{} constructs high-quality benchmarks.}
Human and LLM evaluations both report 100.0 for Evidence Correspondence on 9 of 16 benchmarks and for Task Alignment on 7 of 16.
Per-dimension means are 98.2--99.6\% for human review and 98.3--99.9 out of~100 for LLM judges.
Shading marks values reported as 100.0.}
\label{table:benchmark_quality}
\vspace{-6pt}
\fontsize{8.5pt}{10pt}\selectfont
\setlength{\tabcolsep}{3.5pt}
\renewcommand{\arraystretch}{1.05}
\newcommand{\qualityPending}{\textcolor[rgb]{0.52,0.55,0.59}{\textemdash}}
\definecolor{QualityFullScore}{HTML}{E4F3ED}
\newcommand{\qualityFullScore}{\cellcolor{QualityFullScore}100.0}
\newcommand{\qualityLabelShade}{\cellcolor{ResultRowShade}}

\begin{tabular*}{\linewidth}{ll@{\extracolsep{\fill}\hspace{2\tabcolsep}}*{7}{c}}
\toprule[1.05pt]
\multirow{2}{*}[-.6ex]{\textbf{Modality}}
 & \multirow{2}{*}[-.6ex]{\textbf{Benchmark}}
 & \multicolumn{3}{c}{\cellcolor{humanreview}\textbf{Human Review (\%)}}
 & \multicolumn{4}{c}{\cellcolor{llmjudge}\textbf{LLM Judge (/100)}} \\
\cmidrule(lr){3-5} \cmidrule(lr){6-9}
 & & \shortstack{Evidence\\Corr.}
 & \shortstack{Task\\Align.}
 & \shortstack{Item\\Correctness}
 & \shortstack{Format\\Quality}
 & \shortstack{Evidence\\Corr.}
 & \shortstack{Task\\Align.}
 & \shortstack{Answer\\Validity} \\


\midrule
& Visual Answerability & \qualityFullScore & \qualityFullScore & 98.5
  & \qualityFullScore & 99.6 & \qualityFullScore & 99.2 \\
\qualityLabelShade & \qualityLabelShade Textual Misdirection & \qualityFullScore & \qualityFullScore & 99.0
  & \qualityFullScore & \qualityFullScore & 99.8 & \qualityFullScore \\
& Narrative Causality & 96.0 & 98.0 & 97.5
  & 99.2 & 98.8 & 98.3 & 98.0 \\
\qualityLabelShade & \qualityLabelShade Diagnostic Evidence & \qualityFullScore & \qualityFullScore & 97.0
  & \qualityFullScore & 99.1 & 97.6 & 96.4 \\
& Artwork Reasoning & \qualityFullScore & \qualityFullScore & 98.5
  & \qualityFullScore & \qualityFullScore & 99.2 & 98.9 \\
\qualityLabelShade & \qualityLabelShade Geographic Reasoning & 96.0 & 98.0 & 96.5
  & 99.5 & 98.6 & 97.8 & 98.9 \\
\multirow{-7}{*}{\textbf{Image--Text}}
& Animal Reasoning & 98.0 & 98.0 & 96.0
  & \qualityFullScore & 99.8 & 99.2 & 97.5 \\

\midrule
& Target-Speaker Tracking & \qualityFullScore & \qualityFullScore & \qualityFullScore
  & \qualityFullScore & \qualityFullScore & \qualityFullScore & \qualityFullScore \\
\qualityLabelShade & \qualityLabelShade Speaker Attribution & \qualityFullScore & \qualityFullScore & \qualityFullScore
  & \qualityFullScore & \qualityFullScore & \qualityFullScore & \qualityFullScore \\
& Audio Misdirection & \qualityFullScore & \qualityFullScore & 98.5
  & \qualityFullScore & \qualityFullScore & 97.8 & \qualityFullScore \\
\qualityLabelShade & \qualityLabelShade Audio Answerability & \qualityFullScore & \qualityFullScore & 98.0
  & \qualityFullScore & \qualityFullScore & 99.5 & \qualityFullScore \\
\multirow{-5}{*}{\textbf{Audio--Text}}
& Lyric Interference & \qualityFullScore & \qualityFullScore & \qualityFullScore
  & \qualityFullScore & \qualityFullScore & \qualityFullScore & \qualityFullScore \\

\midrule
& Visual Coverage & 98.0 & \qualityFullScore & 98.5
  & 99.6 & 99.6 & \qualityFullScore & \qualityFullScore \\
\qualityLabelShade & \qualityLabelShade Visual Detail QA & \qualityFullScore & \qualityFullScore & 96.0
  & \qualityFullScore & \qualityFullScore & \qualityFullScore & 90.1 \\
& Scene Misdirection & \qualityFullScore & \qualityFullScore & 98.0
  & \qualityFullScore & \qualityFullScore & \qualityFullScore & 97.9 \\
\qualityLabelShade\multirow{-4}{*}{\textbf{Image--Audio--Text}}
& \qualityLabelShade Art-Historical Reasoning & 98.0 & \qualityFullScore & 98.5
  & 99.8 & 99.4 & 99.8 & 95.5 \\

\bottomrule[1.05pt]
\end{tabular*}
\vspace{-2pt}
\end{table*}

\begin{table*}[t]
\centering
\caption{\textbf{Model performance (\%) on the seven constructed Image--Text benchmarks.}
Each model is evaluated on all 200 items per benchmark. Most open-weight models (8 of 9) average below 65\%. Across open-weight families, the highest scores span 22.5–50.0 percentage points per benchmark. Bold marks column maxima within families with multiple models.}
\label{table:challenge_text_image}
\vspace{-6pt}
\footnotesize
\setlength{\tabcolsep}{2pt}
\renewcommand{\arraystretch}{1.00}
\setlength{\aboverulesep}{1.5pt}
\setlength{\belowrulesep}{1.5pt}

\newcommand{\modelScorePending}{\textcolor[rgb]{0.52,0.55,0.59}{\textemdash}}
\newcommand{\modelScoreID}[1]{\shortstack{#1}}

\definecolor{ModelFamilyBlueRow}{HTML}{F2F8FF}
\definecolor{ModelFamilyGreenRow}{HTML}{F3F9F2}

\begin{tabular*}{\linewidth}{@{\extracolsep{\fill}}ll*{7}{c}@{}}
\toprule[1.05pt]

\multirow{2}{*}{\textbf{Family}}
& \multirow{2}{*}{\textbf{Model}}
& \multicolumn{7}{>{\columncolor{ModelFamilyBlueRow}[0pt][0pt]}c}{\textbf{Image-Text}} \\

\cmidrule(lr){3-9}

& & \modelScoreID{Visual\\Answerability}
& \modelScoreID{Textual\\Misdirection}
& \modelScoreID{Narrative\\Causality}
& \modelScoreID{Diagnostic\\Evidence}
& \modelScoreID{Artwork\\Reasoning}
& \modelScoreID{Geographic\\Reasoning}
& \modelScoreID{Animal\\Reasoning} \\

\midrule
\rowcolor{ModelComparisonGroupRow}[0pt][0pt]
\multicolumn{9}{@{}l}{\textit{\textbf{Proprietary models}}} \\
\midrule

\benchmarkrowbackground{ModelFamilyBlueRow}
& GPT-6.1-Sol
& \textbf{94.0} & \textbf{94.0} & \textbf{96.5} & \textbf{96.5} & \textbf{91.5} & \textbf{98.5} & \textbf{96.5} \\

\benchmarkrowbackground{ModelFamilyBlueRow}
& GPT-5.6-Sol
& 88.5 & 92.0 & 91.0 & 93.0 & 91.0 & 93.5 & 91.5 \\

\benchmarkrowbackground{ModelFamilyBlueRow}
\multirow{-3}{*}{\textit{GPT}}
& GPT-5.6-Luna
& 58.0 & 59.0 & 70.0 & 78.5 & 76.0 & 83.5 & 82.5 \\

\benchmarkrowbackground{ModelFamilyGreenRow}
& Gemini-3.8-Flash
& \textbf{90.5} & \textbf{92.5} & \textbf{91.0} & \textbf{84.0} & \textbf{89.5} & \textbf{98.5} & \textbf{97.0} \\

\benchmarkrowbackground{ModelFamilyGreenRow}
& Gemini-3.6-Flash
& 77.0 & 84.5 & 87.5 & 79.0 & 84.0 & 98.0 & 95.0 \\

\benchmarkrowbackground{ModelFamilyGreenRow}
\multirow{-3}{*}{\textit{Gemini}}
& Gemini-3.1-Flash-Lite
& 44.0 & 59.5 & 64.0 & 66.5 & 69.5 & 73.0 & 68.0 \\

\benchmarkrowbackground{ModelFamilyBlueRow}
\textit{Qwen} & Qwen3.8-Flash
& 59.5 & 70.0 & 76.0 & 77.5 & 74.5 & 79.0 & 76.0 \\

\midrule
\rowcolor{ModelComparisonGroupRow}[0pt][0pt]
\multicolumn{9}{@{}l}{\textit{\textbf{Open-weight models}}} \\
\midrule

\benchmarkrowbackground{ModelFamilyBlueRow}
& Qwen3-VL-8B
& 33.0 & 23.0 & 55.0 & 58.0 & 44.5 & 45.0 & 48.0 \\

\benchmarkrowbackground{ModelFamilyBlueRow}
& Qwen3-VL-32B
& 40.0 & 35.5 & \textbf{68.0} & 62.5 & 54.0 & 58.5 & 60.0 \\

\benchmarkrowbackground{ModelFamilyBlueRow}
& Qwen3.5-9B
& 39.0 & 26.0 & 49.0 & 55.5 & 41.5 & 44.5 & 50.0 \\

\benchmarkrowbackground{ModelFamilyBlueRow}
& Qwen3.5-27B
& 46.0 & 46.0 & 63.0 & 63.0 & 61.5 & 65.0 & 60.0 \\

\benchmarkrowbackground{ModelFamilyBlueRow}
& Qwen3.6-27B
& 42.5 & 44.0 & \textbf{68.0} & 65.0 & 53.5 & 62.0 & 62.5 \\

\benchmarkrowbackground{ModelFamilyBlueRow}
\multirow{-6}{*}{\textit{Qwen}}
& Qwen3.8-27B
& \textbf{57.5} & \textbf{60.0} & 66.5 & \textbf{68.0} & \textbf{64.0} & \textbf{67.0} & \textbf{63.5} \\


\benchmarkrowbackground{ModelFamilyGreenRow}
& InternVL3.5-14B
& 31.0 & 19.0 & 54.0 & 48.5 & \textbf{51.5} & 51.5 & 54.0 \\

\benchmarkrowbackground{ModelFamilyGreenRow}
\multirow{-2}{*}{\textit{InternVL}}
& InternVL3.5-30B-A3B
& \textbf{34.0} & \textbf{21.5} & \textbf{57.5} & \textbf{54.5} & 47.0 & \textbf{53.0} & \textbf{55.0} \\

\benchmarkrowbackground{ModelFamilyBlueRow}
\textit{DeepSeek} & DeepSeek-V4.1-Flash
& 56.0 & 71.5 & 84.5 & 77.0 & 79.0 & 92.5 & 87.0 \\

\bottomrule[1.05pt]
\end{tabular*}
\vspace{-4pt}
\end{table*}

\subsection{Experimental Setup}
\label{sec:experimental_setup}

\noindent\textbf{Benchmark construction.}
We construct 16 benchmarks spanning text, image and audio modalities. Each benchmark is specified by a natural-language evaluation requirement and contains 200 items. We use GPT-6.1-Sol~\cite{openai2026gpt61sol} as the construction backbone and General-Bench~\cite{fei2025path} as the default source pool.
Benchmark details and refinement settings are provided in Appendices~\ref{supp:benchmarks} and~\ref{supp:main_experiments}.

\noindent\textbf{Quality evaluation.}
We assess benchmark quality through human review and LLM-as-a-judge. For each benchmark, two reviewers independently label 50 items as Pass or Fail on \emph{Evidence Correspondence} and \emph{Task Alignment}. An item is counted as Fail on a dimension if either reviewer labels it as Fail. Reviewers also verify the reference answers for all items with GPT-6.1-Sol assistance to assess \emph{Item Correctness}. For LLM-as-a-judge, GPT-6.1-Sol and Seed-2.1-Turbo~\cite{seed2026seed21} independently rate all items from 1 to 5 on these two dimensions, as well as \emph{Format Quality} and \emph{Answer Validity}. Scores are averaged across items and judges and reported on a 100-point scale. Human and LLM evaluation protocols are detailed in Appendices~\ref{supp:human_evaluation} and~\ref{sec:llm as judge}.

\noindent\textbf{Utility evaluation.} We evaluate each benchmark using open-weight and proprietary models from different families, including GPT~\cite{openai2026gpt56,openai2026gpt61sol}, Gemini~\cite{google2026gemini31flashlite,google2026gemini36flash,google2026gemini38flash}, DeepSeek~\cite{xu2026deepseek}, Qwen~\cite{bai2025qwen3vl,qwen3.5,qwen3.6-27b,qwen38,qwen2026design,xu2025qwen3omni,team2026qwen35Omni,team2026qwen38omni}, InternVL~\cite{wang2025internvl3}, and MiniCPM-o~\cite{cui2026minicpm}. We measure model performance using accuracy and assess discrimination based on the absolute performance gap between each model pair.

\subsection{Main Results}

\begin{table*}[t]
\centering
\small
\caption{\textbf{Omni model performance (\%) on the five Audio--Text and four Image--Audio--Text benchmarks.}
The evaluation covers five models, with 200 items in each benchmark.}
\vspace{-6pt}
\label{table:challenge_audio}

\setlength{\tabcolsep}{2pt}
\renewcommand{\arraystretch}{0.92}

\newcommand{\modelScorePending}{\textcolor[rgb]{0.52,0.55,0.59}{\textemdash}}
\newcommand{\modelScoreID}[1]{\shortstack{#1}}

\resizebox{\linewidth}{!}{%
\begin{tabular}{@{}l*{9}{c}@{}}
\toprule[1.05pt]

\multirow{2}{*}{\textbf{Model}}
& \multicolumn{5}{>{\columncolor{AudioText}[0pt][0pt]}c}{\textbf{Audio-Text}}
& \multicolumn{4}{>{\columncolor{TriModal}[0pt][0pt]}c}{\textbf{Image-Audio-Text}} \\

\cmidrule(lr){2-6} \cmidrule(lr){7-10}

& \modelScoreID{Target-Speaker\\Tracking}
& \modelScoreID{{Speaker}\\{Attribution}}
& \modelScoreID{{Audio}\\{Misdirection}}
& \modelScoreID{{Audio}\\{Answerability}}
& \modelScoreID{{Lyric}\\{Interference}}
& \modelScoreID{{Visual}\\{Coverage}}
& \modelScoreID{{Visual}\\{Detail QA}}
& \modelScoreID{{Scene}\\{Misdirection}}
& \modelScoreID{{Art-Historical}\\{Reasoning}} \\

\midrule
\rowcolor{ModelComparisonGroupRow}[0pt][0pt]
\multicolumn{10}{l}{\textit{\textbf{Open-weight models}}} \\
\addlinespace[2pt]
\midrule

MiniCPM-o 4.5
& 38.0 & 35.0 & 31.5 & 32.5 & 33.5
& 49.0 & 41.0 & 1.5 & 39.5 \\

\benchmarkrowbackground{ResultRowShade}
Qwen3-Omni-30B-A3B
& 26.0 & 28.5 & 29.5 & 24.0 & 25.0
& 33.0 & 35.5 & 1.5 & 30.0 \\

Qwen3.5-Omni-Flash
& 35.0 & 33.0 & 30.0 & 27.0 & 48.5
& 41.5 & 53.0 & 21.0 & 49.5 \\

\benchmarkrowbackground{ResultRowShade}
Qwen3.5-Omni-Plus
& 46.5 & 52.5 & 31.5 & 41.0 & 57.5
& 49.0 & 62.5 & 39.5 & 55.0 \\

Qwen3.8-Omni-Flash
& \textbf{52.0} & \textbf{61.0} & \textbf{71.0} & \textbf{61.5} & \textbf{70.5}
& \textbf{66.0} & \textbf{69.0} & \textbf{52.0} & \textbf{63.5} \\

\bottomrule[1.05pt]
\end{tabular}%
}
\vspace{-6pt}
\end{table*}

\begin{figure*}[!t]
    \centering
    \vspace{4pt}
    \includegraphics[width=\linewidth]{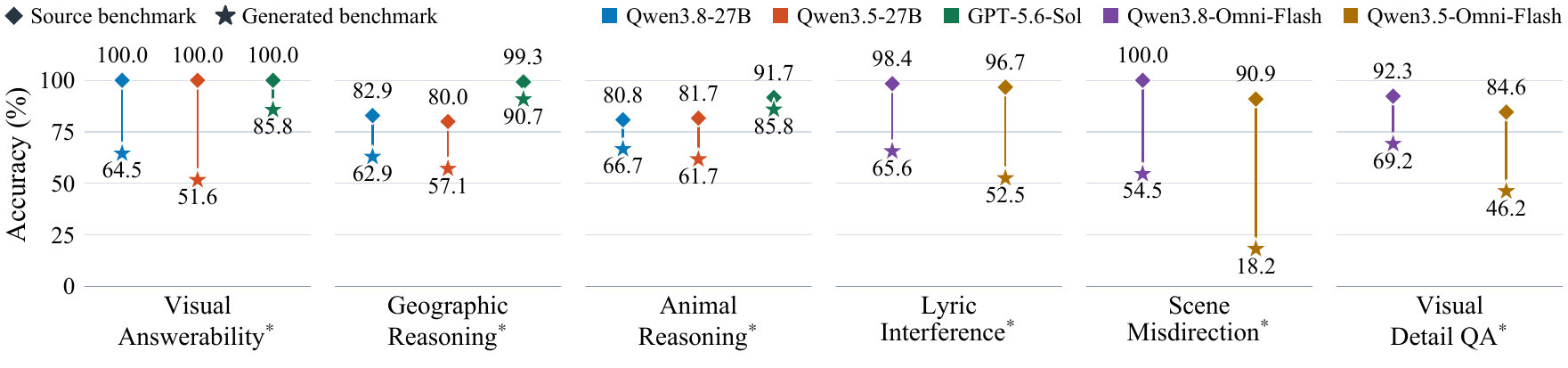}
    \vspace{-12pt}
\caption{\textbf{Benchmark difficulty.}
Lines connect each model's scores on source samples (diamonds) and generated items (stars). `*’ mark the subsets used for the paired comparison.}
    \label{fig:difficulty}
\end{figure*}

\begin{figure}[!t]
    \centering
    \vspace{-8pt}
    \includegraphics[width=\linewidth]{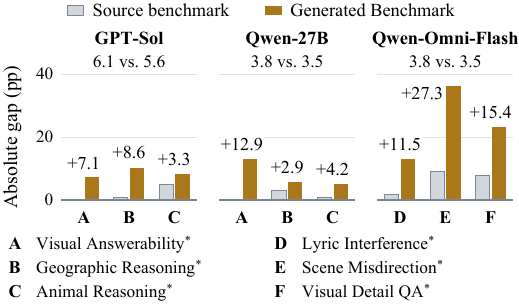}
        \vspace{-6pt}
\caption{\textbf{Benchmark discrimination.}
Bars show absolute performance gaps within each model pair. Labels give the change from source to generated items in percentage points. `*' marks the subsets used for the paired comparison.}
    \label{fig:discrimination}
    \vspace{-4pt}
\end{figure}

\begin{figure*}[!t]
    \centering
    \includegraphics[width=0.93\linewidth]{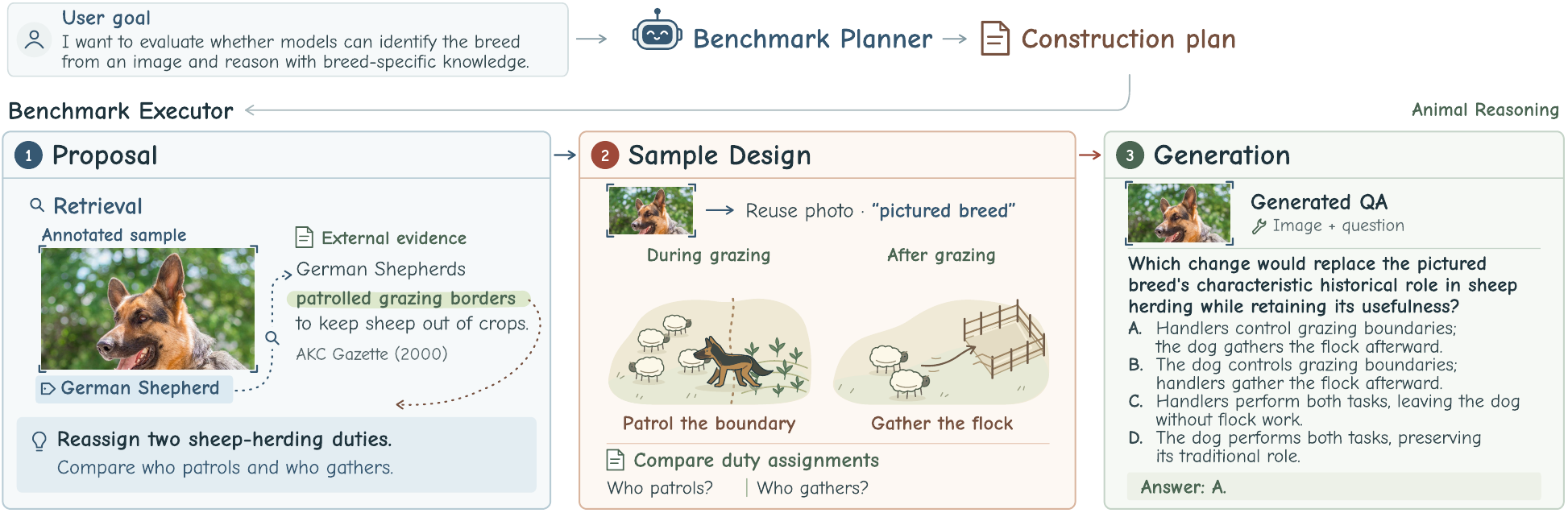}
\caption{\textbf{Animal Reasoning example.}
\methodname{} asks how to reassign a German Shepherd's herding duties.}
    \label{fig:case_study}
    \vspace{-10pt}
\end{figure*}

\noindent\textbf{Benchmark quality.} As shown in Table~\ref{table:benchmark_quality}, human review yields average pass rates of 99.1\% and 99.6\% for Evidence Correspondence and Task Alignment, with 98.2\% Item Correctness. LLM judge scores average 98.3--99.9 across the four dimensions. These results indicate that \methodname{} constructs benchmarks that align with their intended tasks and contain reliable evidence and answers.

We further assess the consistency of the quality evaluation by comparing human and language model judgments (Figure~\ref{fig:judge_agreement}). As most items receive Pass or 5, we focus on cases with lower assessments. Among items marked as Fail by either human reviewer, most are marked as Fail by both, and most receive a score of 1 from both model judges. When at least one model judge scores an item below 5, their scores usually fall in the same range (3--4 or 1--2). This agreement supports the reliability of our quality evaluation.

\noindent\textbf{Benchmark Utility.}
We evaluate models of different families, sizes, and versions on the constructed benchmarks (Tables~\ref{table:challenge_text_image} and~\ref{table:challenge_audio}). \textit{(i) Benchmark difficulty.} Eight of the nine open-weight models average below 65\% across the seven Image--Text benchmarks. All five evaluated Omni models also average below 65\% across the nine audio-related benchmarks, with the highest mean of 62.9\% achieved by Qwen3.8-Omni-Flash. \textit{(ii) Model discrimination.} On all seven Image--Text benchmarks, Qwen3-VL-32B and Qwen3.5-27B outperform their smaller counterparts. The four Qwen Omni models likewise maintain the same ranking across all nine audio-related benchmarks. Differences remain visible among strong proprietary models: GPT-6.1-Sol outperforms GPT-5.6-Sol, and Gemini-3.8-Flash outperforms Gemini-3.6-Flash. These results suggest that the benchmarks challenge a range of models and reveal differences across model sizes and versions. Selected model failures are illustrated in Appendix~\ref{supp:benchmark_visualization}.

To examine the utility of benchmarks constructed from existing data, we select six benchmarks from Table~\ref{table:benchmark_quality}  whose source datasets have original tasks that can be evaluated using accuracy.  We test the same models on the generated items and their corresponding source samples.
As shown in Fig.~\ref{fig:difficulty}, models achieve 80.0--100.0\% on the original samples, but their performance decreases by 5.8--72.7  percentage points on the generated items.
The performance gaps between model pairs also increase by 2.9--27.3 percentage points (Fig.~\ref{fig:discrimination}).
These results show that \methodname{} can construct challenging benchmarks that distinguish models from existing data, even when performance on the original tasks approaches saturation.

\begin{table}[!b]
\centering
\caption{\textbf{Direct Generation vs.\ \methodname{} across backbones.}
Pairs show \textit{Direct Generation} $\to$ \textit{\methodname{}}.}
\vspace{-5pt}
\label{table:generation_backbone}
\label{table:direct_llm}
\label{table:agent_base_model_sensitivity}

\fontsize{7.5pt}{9pt}\selectfont
\setlength{\tabcolsep}{1.2pt}
\renewcommand{\arraystretch}{1.10}

\definecolor{ScoreArrowGray}{RGB}{88,96,105}

\newcommand{\scoreArrow}{%
  \ensuremath{%
    \mkern1.2mu
    \textcolor{ScoreArrowGray}{\boldsymbol{\to}}%
    \mkern1.2mu
  }%
}
\newcommand{\backbonePair}[2]{#1\scoreArrow#2}
\newcommand{\backboneMetric}[1]{\shortstack{#1}}

\resizebox{\linewidth}{!}{%
\begin{tabular}{@{}l*{4}{c}@{}}
\toprule[1.05pt]
\multirow{2}{*}{\textbf{Backbone}}
 & \multicolumn{4}{>{\columncolor[rgb]{0.91,0.96,0.94}[0pt][0pt]}c}
     {\textbf{LLM Judge}} \\
\cmidrule(lr){2-5}
 & \backboneMetric{Format\\Quality}
 & \backboneMetric{Evidence\\Corr.}
 & \backboneMetric{Task\\Align.}
 & \backboneMetric{Answer\\Validity} \\
\midrule

\rowcolor{BenchmarkTitleRow}[0pt][0pt]
\multicolumn{5}{@{}l}{\textit{Artwork Reasoning}} \\[1pt]
\cc GPT-6.1-Sol
 & \backbonePair{100.0}{100.0}
 & \backbonePair{97.5}{100.0}
 & \backbonePair{89.2}{99.2}
 & \backbonePair{98.7}{98.9} \\
\cc Gemini-3.8-Flash
 & \backbonePair{100.0}{100.0}
 & \backbonePair{98.0}{99.8}
 & \backbonePair{74.4}{86.3}
 & \backbonePair{90.1}{95.2} \\
\oc DeepSeek-V4.1-Flash
 & \backbonePair{100.0}{100.0}
 & \backbonePair{95.8}{96.3}
 & \backbonePair{86.3}{94.5}
 & \backbonePair{85.8}{91.2} \\
\oc Qwen3.5-397B-A17B
 & \backbonePair{100.0}{100.0}
 & \backbonePair{96.5}{99.6}
 & \backbonePair{83.2}{94.1}
 & \backbonePair{83.2}{95.7} \\
\midrule

\rowcolor{BenchmarkTitleRow}[0pt][0pt]
\multicolumn{5}{@{}l}{\textit{Audio Answerability}} \\[1pt]
\cc GPT-6.1-Sol
 & \backbonePair{99.0}{100.0}
 & \backbonePair{98.8}{100.0}
 & \backbonePair{91.3}{99.5}
 & \backbonePair{99.6}{100.0} \\
\cc Gemini-3.8-Flash
 & \backbonePair{100.0}{100.0}
 & \backbonePair{100.0}{100.0}
 & \backbonePair{64.5}{85.4}
 & \backbonePair{91.8}{94.4} \\
\oc DeepSeek-V4.1-Flash
 & \backbonePair{99.8}{99.8}
 & \backbonePair{100.0}{100.0}
 & \backbonePair{61.4}{89.2}
 & \backbonePair{95.9}{98.6} \\
\oc Qwen3.5-397B-A17B
 & \backbonePair{85.4}{100.0}
 & \backbonePair{81.7}{98.4}
 & \backbonePair{73.3}{94.8}
 & \backbonePair{82.2}{90.9} \\
\midrule

\rowcolor{BenchmarkTitleRow}[0pt][0pt]
\multicolumn{5}{@{}l}{\textit{Scene Misdirection}} \\[1pt]
\cc GPT-6.1-Sol
 & \backbonePair{100.0}{100.0}
 & \backbonePair{100.0}{100.0}
 & \backbonePair{89.8}{100.0}
 & \backbonePair{95.9}{97.9} \\
\cc Gemini-3.8-Flash
 & \backbonePair{100.0}{100.0}
 & \backbonePair{99.8}{100.0}
 & \backbonePair{77.5}{86.6}
 & \backbonePair{88.1}{93.4} \\
\oc DeepSeek-V4.1-Flash
 & \backbonePair{99.8}{100.0}
 & \backbonePair{99.1}{100.0}
 & \backbonePair{78.6}{88.6}
 & \backbonePair{86.4}{94.7} \\
\oc Qwen3.5-397B-A17B
 & \backbonePair{99.6}{99.6}
 & \backbonePair{92.4}{93.6}
 & \backbonePair{68.8}{87.2}
 & \backbonePair{96.1}{98.1} \\
\bottomrule[1.05pt]
\end{tabular}%
}
\end{table}

\noindent\textbf{Qualitative example.}
Figure~\ref{fig:case_study} illustrates the construction of an item for an animal reasoning requirement. \methodname{} selects an image of a German Shepherd and its breed label from a retrieved dataset. Guided by the requirement, it retrieves evidence on the breed's historical role in maintaining grazing boundaries and uses this evidence to design a scenario in which the dog and human handlers take on different herding duties. The resulting item retains the image and asks which reassignment would replace the breed's characteristic role while leaving the dog with useful flock work.

\subsection{Ablation Studies}\label{sec:ablation}

\begin{table}[t]
\centering
\caption{\textbf{Component ablation across three benchmarks.}
Final Review pass rate is the percentage of submitted candidates accepted
by the shared review stage.}
\vspace{-5pt}
\label{tab:component_ablation}

\fontsize{8pt}{9.5pt}\selectfont
\setlength{\tabcolsep}{3.3pt}
\renewcommand{\arraystretch}{1.05}

\definecolor{AblationFullRow}{RGB}{240,242,244}

\newcommand{\ablationHead}[1]{%
  \multicolumn{1}{c}{\shortstack{#1}}%
}

\begin{tabular*}{\columnwidth}{@{\extracolsep{\fill}}l*{5}{r}@{}}
\toprule[1.05pt]
\multirow{2}{*}[-0.5ex]{\textbf{Setting}}
 & \multicolumn{4}{>{\columncolor[RGB]{237,249,242}[0pt][0pt]}c}
     {\textbf{LLM Judge}}
 & \multicolumn{1}{>{\columncolor[RGB]{245,242,249}[0pt][0pt]}c}
     {\textbf{Review}} \\
\cmidrule(lr){2-5} \cmidrule(lr){6-6}
 & \ablationHead{Format\\Quality}
 & \ablationHead{Evidence\\Corr.}
 & \ablationHead{Task\\Align.}
 & \ablationHead{Answer\\Validity}
 & \ablationHead{Pass\\(\%)} \\
\midrule
\addlinespace[1pt]

\benchmarkrowbackground{BenchmarkTitleRow}
\multicolumn{6}{l}{\textit{Artwork Reasoning}} \\
\addlinespace[1.5pt]
\benchmarkrowbackground{AblationFullRow}
\textbf{\methodname{}}
 & 100.0 & 100.0 & 99.2 & 98.9 & 72.0 \\
\hspace{3pt}w/o Design
 & 100.0 & 98.9 & 95.8 & 98.1 & 47.5 \\
\benchmarkrowbackground{ResultRowShade}
\hspace{3pt}w/o Grounding
 & 100.0 & 100.0 & 95.8 & 96.8 & 26.5 \\
\hspace{3pt}w/o Executor
 & 100.0 & 99.2 & 94.2 & 98.2 & 15.5 \\
\benchmarkrowbackground{ResultRowShade}
\hspace{3pt}w/o Refinement
 & 100.0 & 100.0 & 99.6 & 99.0 & 62.5 \\
\midrule
\addlinespace[4pt]
\benchmarkrowbackground{BenchmarkTitleRow}
\multicolumn{6}{l}{\textit{Audio Answerability}} \\
\addlinespace[1.5pt]
\benchmarkrowbackground{AblationFullRow}
\textbf{\methodname{}}
 & 100.0 & 100.0 & 99.5 & 100.0 & 87.0 \\
\hspace{3pt}w/o Design
 & 100.0 & 100.0 & 87.1 & 88.8 & 32.0 \\
\benchmarkrowbackground{ResultRowShade}
\hspace{3pt}w/o Grounding
 & 100.0 & 100.0 & 91.6 & 95.9 & 80.5 \\
\hspace{3pt}w/o Executor
 & 99.3 & 98.9 & 95.9 & 98.0 & 23.0 \\
\benchmarkrowbackground{ResultRowShade}
\hspace{3pt}w/o Refinement
 & 100.0 & 100.0 & 99.7 & 100.0 & 79.5 \\
\midrule
\addlinespace[4pt]
\benchmarkrowbackground{BenchmarkTitleRow}
\multicolumn{6}{l}{\textit{Scene Misdirection}} \\
\addlinespace[1.5pt]
\benchmarkrowbackground{AblationFullRow}
\textbf{\methodname{}}
 & 100.0 & 100.0 & 100.0 & 97.9 & 71.0 \\
\hspace{3pt}w/o Design
 & 100.0 & 100.0 & 68.8 & 93.8 & 43.5 \\
\benchmarkrowbackground{ResultRowShade}
\hspace{3pt}w/o Grounding
 & 100.0 & 97.6 & 94.0 & 91.1 & 52.5 \\
\hspace{3pt}w/o Executor
 & 99.3 & 99.0 & 80.3 & 90.0 & 18.0 \\
\benchmarkrowbackground{ResultRowShade}
\hspace{3pt}w/o Refinement
 & 100.0 & 100.0 & 99.6 & 99.4 & 61.0 \\
\bottomrule[1.05pt]
\end{tabular*}
\vspace{-10pt}
\end{table}
We conduct the following ablation studies on Artwork Reasoning, Audio Answerability, and Scene Misdirection, using the settings in Appendix~\ref{supp:ablation_studies}.

\noindent\textbf{Effect of the harness across backbones.}
\label{sec:baseline}
\label{sec:ablation_backbone}
We compare Direct Generation and \methodname{} using four LLM backbones. Both methods receive the same requirement, produce 200 items, and share the same transformation tool library. Direct Generation uses a single LLM-based agent for dataset retrieval, item construction, and review (Appendix~\ref{supp:direct_generation}). As shown in Table~\ref{table:generation_backbone}, \methodname{} improves benchmark quality across all backbones: average Task Alignment rises from 78.2 to 92.1, and Answer Validity from 91.2 to 95.8. All judge scores with \methodname{} are at least 85.4. These results demonstrate the effectiveness of \methodname{} in improving benchmark quality across LLM backbones.

\noindent\textbf{Component ablation.}\label{sec:ablation_components}
We fix the backbone to GPT-6.1-Sol and ablate Design, Grounding, Executor, and Refinement on the same three benchmarks (Table~\ref{tab:component_ablation}). The variants respectively use a single task, basic dataset search, a lightweight execution agent with the shared tools, and no refinement (Appendix~\ref{supp:component_ablation}). The full system achieves an average Final Review pass rate of 76.7\%, compared with 18.8--67.7\% for the variants. It also receives higher average Task Alignment and Answer Validity scores than the first three variants, while w/o Refinement achieves comparable judge scores. These results highlight the contribution of the components to benchmark construction.

\section{Discussion}
\methodname{} repurposes annotated data into benchmarks for user-defined tasks.
Across modalities, the resulting items align with those tasks and distinguish models.
Model feedback can guide further refinement.. Across tasks and modalities, our evaluations show that the \methodname{} produces well-supported items aligned with their intended tasks and that the resulting benchmarks are challenging and discriminative.
As Fig.~\ref{fig:motivation} illustrates, fixed benchmarks can become less informative as models improve. Evaluation needs also vary across domains and applications. \methodname{} points to a new direction for model evaluation by making benchmark construction a reusable process guided by evaluation goals. The same workflow can turn new questions about model capabilities into benchmarks, with existing annotated datasets providing the material for new tasks as models and their applications evolve. Limitations are discussed in Appendix~\ref{supp:limitations}.


\FloatBarrier
\setlength{\textfloatsep}{\maintextfloatsep}
\flushbottom
{
    \small
    \bibliographystyle{ieeenat_fullname}
    \bibliography{main}
}

\clearpage
\appendix

\raggedbottom
\makeatletter
\setlength{\@fptop}{0pt}
\setlength{\@dblfptop}{0pt}
\makeatother

\renewcommand\thefigure{\Alph{section}\arabic{figure}}
\renewcommand\thetable{\Alph{section}\arabic{table}} 

\section*{Contents}

\startcontents[appendices] 
\setcounter{tocdepth}{3} 
\printcontents[appendices]{}{1}{} 

\newtcolorbox{ratingbox}[1]{
    enhanced,breakable,arc=1mm,boxrule=0.5pt,
    colback=white,colframe=black!25,
    colbacktitle=black!7,coltitle=black,
    title={#1},title after break={#1\space(continued)},
    fonttitle=\rmfamily\bfseries,fontupper=\rmfamily,
    left=8pt,right=8pt,top=4pt,bottom=4pt,boxsep=1.5pt,
    before skip=7pt,after skip=7pt,
    before upper={\setlength{\parindent}{0pt}\setlength{\parskip}{3pt}}
}

\newpage
\section{Benchmark Details}

\subsection{Benchmark Overview}
\label{supp:benchmarks}

\begingroup
\small
\rmfamily
\definecolor{BenchInk}{HTML}{293B49}
\definecolor{BenchMuted}{HTML}{667783}
\definecolor{BenchStripe}{HTML}{F7F9FB}
\definecolor{BenchRule}{HTML}{D9E1E7}
\definecolor{BenchImage}{HTML}{67517D}
\definecolor{BenchAudio}{HTML}{355E78}
\definecolor{BenchOmni}{HTML}{8A6138}
\setlength{\tabcolsep}{4pt}
\setlength{\extrarowheight}{2pt}
\setlength{\parskip}{0pt}
\renewcommand{\arraystretch}{1.12}
\newcommand{\benchhead}[1]{\rule[-3pt]{0pt}{16pt}{\rmfamily\bfseries\color{white}#1}}
\newcommand{\benchsubtask}[1]{%
    \par\noindent\hangindent=0.85em\hangafter=1%
    \makebox[0.85em][l]{\textcolor{BenchMuted}{\raisebox{0.15ex}{\scriptsize\textbullet}}}#1%
}
\newcommand{\benchgroup}[4]{%
    \rowcolor{#1}%
    \multicolumn{3}{p{\dimexpr\linewidth-2\tabcolsep\relax}}{%
        \rule[-4pt]{0pt}{18pt}%
        {\rmfamily\bfseries\color{#2}#3}\hfill%
        {\rmfamily\footnotesize\color{#2}#4 benchmarks}%
    }\\*
}
\begin{table*}[!t]
\centering\small\rmfamily
\caption{User requirements and subtasks for the 16 benchmarks in Table~\ref{table:benchmark_quality}, grouped by input modality.}
\label{supp_table:benchmark_overview}
\begin{tabular}{
    >{\raggedright\arraybackslash}p{\dimexpr0.22\linewidth-2\tabcolsep\relax}
    >{\raggedright\arraybackslash}p{\dimexpr0.49\linewidth-2\tabcolsep\relax}
    >{\raggedright\arraybackslash}p{\dimexpr0.29\linewidth-2\tabcolsep\relax}
}
\arrayrulecolor{BenchInk}\toprule
\rowcolor{BenchInk}
\benchhead{Benchmark} & \benchhead{User Requirement} & \benchhead{Subtasks} \\
\arrayrulecolor{BenchRule}\midrule
\benchgroup{ImageText}{BenchImage}{Image + Text}{7}
\rowcolor{white}
{\bfseries\color{BenchInk} Artwork Reasoning} &
I want to build a benchmark where models infer an unlabeled artwork’s identity, artist, or period from the image, then reason over related art history, biography, or historical context. Questions should require knowledge beyond what is directly visible. &
\benchsubtask{Artwork-grounded artist biography reasoning}
\benchsubtask{Artwork-grounded artistic relationships and oeuvre reasoning}
\benchsubtask{Artwork-grounded historical and cultural reasoning} \\[3pt]
\rowcolor{BenchStripe}
{\bfseries\color{BenchInk} Textual Misdirection} &
I want to assess whether models can verify text that mixes accurate observations with plausible but misleading interpretations against an image. They should retain supported observations and reject misleading conclusions by integrating the visual evidence. &
\benchsubtask{Visually grounded judgment under misleading interpretation}
\benchsubtask{Selective acceptance of mixed visual claims} \\[3pt]
\rowcolor{white}
{\bfseries\color{BenchInk} Visual Answerability} &
I want a benchmark assessing whether models answer accurately when visual evidence is sufficient and abstain when information is missing or ambiguous. It should distinguish confirmed absence from insufficient evidence, avoiding unsupported answers and unnecessary abstention. &
\benchsubtask{Evidence-calibrated visual question answering}
\benchsubtask{Confirmed absence versus unresolved presence} \\[3pt]
\rowcolor{BenchStripe}
{\bfseries\color{BenchInk} Geographic Reasoning} &
I want to evaluate whether models can infer a place, landform, or region from an image without being given its identity. They should combine entity-specific knowledge, geographic relationships, and the question’s conditions to reach justified conclusions. &
\benchsubtask{Visually grounded geographic explanation}
\benchsubtask{Visual gateway to geographic relationship chains}
\benchsubtask{Conditional geographic consequence inference} \\[3pt]
\rowcolor{white}
{\bfseries\color{BenchInk} Diagnostic Evidence} &
I want to examine whether models can combine medical images, clinical context, and medical knowledge to compare competing explanations. They should select evidence that distinguishes those explanations and update conclusions after further observations or tests, accounting for uncertainty and test limitations. &
\benchsubtask{Image-grounded selection of discriminating evidence}
\benchsubtask{Image-grounded differential updating after further evidence} \\[3pt]
\rowcolor{BenchStripe}
{\bfseries\color{BenchInk} Animal Reasoning} &
I want a benchmark that assesses whether models can infer an animal’s breed or species from visual evidence and apply relevant category knowledge. Questions should connect breed origins and historical roles, or species habits, life history, and ecology, with the stated conditions. &
\benchsubtask{Visually anchored breed-history reasoning}
\benchsubtask{Visually anchored species-ecology reasoning} \\[3pt]
\rowcolor{white}
{\bfseries\color{BenchInk} Narrative Causality} &
I want to evaluate models’ ability to integrate visual evidence, dialogue, and character knowledge across panels to infer omitted events and judge causal explanations or continuations. Judgments should remain consistent with earlier evidence and what each character knows. &
\benchsubtask{Cross-panel omitted-event reconstruction}
\benchsubtask{Narrative-wide causal account and continuation assessment} \\[3pt]
\benchgroup{AudioText}{BenchAudio}{Audio + Text}{5}
\rowcolor{white}
{\bfseries\color{BenchInk} Audio Misdirection} &
I want to build a speech benchmark where models resolve persuasive textual distortions of reference, scope, causality, conditions, or exceptions by integrating separated audio passages. Audio should remain intelligible and authoritative; isolated statements or background knowledge should not suffice. &
\benchsubtask{Audio-grounded reference and scope reconstruction}
\benchsubtask{Audio-grounded causal and conditional interpretation} \\[3pt]
\arrayrulecolor{BenchInk}\bottomrule
\end{tabular}
\end{table*}

\begin{table*}[!t]
\centering\small\rmfamily
\ContinuedFloat
\caption[]{User requirements and subtasks (continued).}
\begin{tabular}{
    >{\raggedright\arraybackslash}p{\dimexpr0.22\linewidth-2\tabcolsep\relax}
    >{\raggedright\arraybackslash}p{\dimexpr0.49\linewidth-2\tabcolsep\relax}
    >{\raggedright\arraybackslash}p{\dimexpr0.29\linewidth-2\tabcolsep\relax}
}
\arrayrulecolor{BenchInk}\toprule
\rowcolor{BenchInk}
\benchhead{Benchmark} & \benchhead{User Requirement} & \benchhead{Subtasks} \\
\arrayrulecolor{BenchRule}\midrule
\benchgroup{AudioText}{BenchAudio}{Audio + Text (continued)}{5}
\rowcolor{BenchStripe}
{\bfseries\color{BenchInk} Target-Speaker Tracking} &
I want to measure whether models can use a reference voice to recover target statements during overlapping speech and combine them with context to reason about facts, relationships, or constraints. Later speech should not clearly repeat the answer-critical information. &
\benchsubtask{Target-attributed fact recovery in context}
\benchsubtask{Target-grounded relationship and constraint reasoning} \\[3pt]
\rowcolor{white}
{\bfseries\color{BenchInk} Audio Answerability} &
I want a benchmark for judging yes, no, or insufficient information from dispersed speech. It should require resolving references, temporal and reporting coverage, and sources, while distinguishing confirmed absence from missing evidence and attributed claims from verified facts without wording cues. &
\benchsubtask{Temporal and reporting coverage sufficiency}
\benchsubtask{Attribution and source-grounded answerability} \\[3pt]
\rowcolor{BenchStripe}
{\bfseries\color{BenchInk} Lyric Interference} &
I want to evaluate whether models can integrate spoken facts, conditions, negations, and corrections while ignoring sung lyrics overlapping critical speech. Interference should vary in strength, with enough intelligibility for a careful listener to recover the required evidence. &
\benchsubtask{Resolving spoken conditions, negations, and corrections} \\[3pt]
\rowcolor{white}
{\bfseries\color{BenchInk} Speaker Attribution} &
I want a benchmark for binding a target speaker’s facts, conditions, and revisions across nonadjacent turns using a reference voice. Speakers should discuss similar entities under continuous music, without overlapping speech or later recaps that reveal the attribution. &
\benchsubtask{Reference-voice cross-turn information binding}
\benchsubtask{Reference-voice condition and revision binding} \\[3pt]
\benchgroup{TriModal}{BenchOmni}{Image + Audio + Text}{4}
\rowcolor{white}
{\bfseries\color{BenchInk} Visual Coverage} &
I want to assess whether models can judge presence, absence, or “cannot determine” from inspection scope, visibility, occlusion, resolution, and complementary views. All three outcomes should be represented, and speech must not reveal unseen facts that settle the answer. &
\benchsubtask{Scoped inspection judgment from one photograph}
\benchsubtask{Joint inspection coverage across photographs} \\[3pt]
\rowcolor{BenchStripe}
{\bfseries\color{BenchInk} Scene Misdirection} &
I want to test whether models can answer from photographs when speech and text reinforce the same plausible but incorrect interpretation. Objects, parts, actions, or relationships should admit plausible competing readings, with visual evidence allowing models to resolve the conflict. &
\benchsubtask{Object identity and visual ownership under persuasive commentary}
\benchsubtask{Action, relationship, and scope under persuasive commentary} \\[3pt]
\rowcolor{white}
{\bfseries\color{BenchInk} Visual Detail QA} &
I want to assess visual-detail answerability using actual visibility and delivered resolution: answer when evidence suffices and choose “cannot determine” otherwise. Speech must not reveal hidden details or declare insufficiency, and wording must not signal uncertainty. &
\benchsubtask{Single-photo evidence-grounded detail answering}
\benchsubtask{Joint evidence grounding across photographic views} \\[3pt]
\rowcolor{BenchStripe}
{\bfseries\color{BenchInk} Art-Historical Reasoning} &
I want a benchmark where models infer an artwork, creator, or historical setting from an unlabeled image, then reason about relevant artists, workshops, patrons, commissions, or other works. Speech and text must not reveal the visual identity or tested relationship. &
\benchsubtask{Artwork-grounded relationship reasoning with exhibition speech} \\[3pt]
\arrayrulecolor{BenchInk}\bottomrule
\end{tabular}
\end{table*}

\endgroup

Table~\ref{supp_table:benchmark_overview} summarizes the 16 benchmarks constructed with \methodname{}, grouped by input modality and listing their user requirements and subtasks. Seven benchmarks use image and text inputs, five use audio and text inputs, and four combine image, audio, and text.

\par\medskip
\section{Implementation Details}
\label{sec:implementation details}


\subsection{Transformation Tools}
\label{supp:transformation_tools}
Sample transformation tools generate, modify, and organize multimodal benchmark samples.

\begin{itemize}
      \item \textbf{Text-to-speech} converts text into independently reusable speech segments with configurable language, speaking rate, volume, and pitch, while keeping each speaker's voice consistent.
    \item \textbf{Speech segment mixing} concatenates segments in sequence or overlaps them at specified start times, with per-segment pause and gain controls.
    \item \textbf{Environmental noise injection} mixes noise from a specified environment into audio at low, medium, or high intensity.
    \item \textbf{Background music mixing} adds music with English vocals and accompaniment at a specified speech-to-music level ratio while preserving the foreground speech duration.
    \item \textbf{Image degradation} applies spatial degradation, photometric distortion, compression artifacts, occlusion, or noise using a selected mode and a severity level from 1 to 5.
    \item \textbf{Speech-to-text} defines an interface for transcribing input speech audio into text.
\end{itemize}

\clearpage
\makeatletter
\global\@colht=\textheight
\global\@colroom=\textheight
\global\vsize=\textheight
\makeatother
\subsection{Prompt Templates}
\label{supp:prompt_templates}
\raggedbottom
\definecolor{PromptInk}{HTML}{293B49}
\definecolor{PromptRule}{HTML}{D8E0E7}
\definecolor{PromptCodeBackground}{HTML}{F7F9FB}
\definecolor{PromptBlue}{HTML}{496A86}
\definecolor{PromptTeal}{HTML}{42766B}
\definecolor{PromptPurple}{HTML}{755B88}
\newcommand{\promptfield}[1]{\textcolor{PromptBlue}{#1}}
\newcommand{\prompthead}[1]{\par\addvspace{3pt}\noindent\textbf{\color{PromptInk}#1}\par\nobreak}
\newtcolorbox{agentprompt}[2]{
    enhanced,breakable,arc=1mm,boxrule=0.5pt,
    colback=white,colframe=#1!45!white,
    colbacktitle=#1!10!white,coltitle=PromptInk,
    borderline west={1.7pt}{0pt}{#1},
    title={#2},title after break={#2\space(continued)},
    fonttitle=\rmfamily\bfseries\normalsize,
    fontupper=\rmfamily\fontsize{9}{10.6}\selectfont,
    fontlower=\rmfamily\fontsize{9}{10.6}\selectfont,
    colupper=PromptInk,collower=PromptInk,
    left=8pt,right=8pt,top=4pt,bottom=4pt,boxsep=1.5pt,
    before skip=7pt,after skip=7pt,
    before upper={\setlength{\parindent}{0pt}\setlength{\parskip}{2.5pt}\raggedright\hyphenpenalty=10000\exhyphenpenalty=10000},
    before lower={\setlength{\parindent}{0pt}\setlength{\parskip}{1.5pt}\raggedright\hyphenpenalty=10000\exhyphenpenalty=10000}
}
\newtcolorbox{promptjson}[2]{
    enhanced,breakable,arc=1mm,boxrule=0.5pt,
    colback=PromptCodeBackground,colframe=PromptRule,
    colbacktitle=#1!7!white,coltitle=PromptInk,
    title={#2},title after break={#2\space(continued)},
    fonttitle=\rmfamily\bfseries\small,
    fontupper=\rmfamily\footnotesize,
    left=8pt,right=8pt,top=3pt,bottom=3pt,boxsep=1pt,
    before skip=6pt,after skip=6pt
}

\subsubsection{Planner Agents}

\label{supp:planner_agent_prompts}

We provide the main prompt templates for the planner's Design, Grounding, and Allocation agents. Braced fields are runtime inputs; angle-bracketed strings are illustrative JSON values.

\begin{agentprompt}{PromptBlue}{Design Agent: Prompt Template}

You are the benchmark Design Agent. Turn the user's evaluation goal into a coherent working set of 1-3 subtasks, ready for Grounding. 
\par
Define each subtask as an individual input-to-answer task; comparisons across samples are outside the subtask requirements.
\par
\prompthead{Design criteria}
\begin{itemize}[leftmargin=1.15em,labelsep=0.4em,topsep=1.5pt,partopsep=0pt,itemsep=1pt,parsep=0pt]
\item Every subtask must directly exercise the user's core capability. Give each a clear evaluation purpose; together the set should cover the requested benchmark.
\item Make descriptions concise: state what capability is tested and the distinction that matters.
\item Use existing subtask fields: id, name, description, answer\_\allowbreak{}type, modalities, sample\_\allowbreak{}schema, keywords. Preserve the user's explicit requirements.
\item Input requires question; add context, image\_\allowbreak{}url or audio\_\allowbreak{}url only when needed. For media tasks avoid redundant textual replacements for the required perception. Output contains only answer. Choose binary, choice, label or span to suit the capability; prefer choice when equally suitable, without weakening the target. Choice options belong in question, not context or a separate options field.
\end{itemize}
\par
\prompthead{Decision and tools}
Use the working state, candidates and feedback to choose the next action. Include a brief decision\_\allowbreak{}rationale in tool calls.
\begin{itemize}[leftmargin=1.15em,labelsep=0.4em,topsep=1.5pt,partopsep=0pt,itemsep=1pt,parsep=0pt]
\item If candidates are pending, resolve them with \textbf{process\_\allowbreak{}proposed\_\allowbreak{}subtasks}(accepted\_\allowbreak{}subtask\_\allowbreak{}ids): accept useful candidates or pass [] to reject them.
\item Call \textbf{proposer}(guidance=...) when the working set is empty or a specific coverage/design gap needs new candidates, with no unresolved proposal.
\item Call \textbf{llm\_\allowbreak{}revise\_\allowbreak{}one\_\allowbreak{}subtask}(subtask\_\allowbreak{}id, guidance) for a specific defect in an accepted task.
\item Call \textbf{discard\_\allowbreak{}subtask} for redundancy, target mismatch or an unresolved flaw that cannot usefully be revised; check remaining coverage.
\item Call \textbf{case\_\allowbreak{}resolved}(summary) when the nonempty working set meets the design criteria, pending candidates are resolved, and no concrete design defect remains.
\end{itemize}
\par
Apply justified local changes in response to feedback. Preserve the core target and unaffected tasks.
\par
\tcblower
\begingroup
\setlength{\parskip}{1.5pt}
\noindent\textbf{User requirement:} \promptfield{\{description\}}\par
\noindent\textbf{Scope context:} short\_\allowbreak{}topic=\promptfield{\{short\_\allowbreak{}topic\}}; modalities=\promptfield{\{modalities\}}; keywords=\promptfield{\{keywords\}}\par
\noindent\textbf{Working subtasks:} \promptfield{\{subtasks\_\allowbreak{}str\}}\par
\noindent\textbf{Pending candidates:} \promptfield{\{proposed\_\allowbreak{}str\}}\par
\noindent\textbf{Earlier Design runs:} \promptfield{\{design\_\allowbreak{}agent\_\allowbreak{}history\_\allowbreak{}str\}}\par
\noindent\textbf{Grounding feedback:} \promptfield{\{grounding\_\allowbreak{}feedback\}}\par
\noindent\textbf{Allocation feedback:} \promptfield{\{allocation\_\allowbreak{}feedback\}}\par
\endgroup

\end{agentprompt}

\begin{promptjson}{PromptBlue}{Example: Subtask}
\begin{lstlisting}[basicstyle=\rmfamily\fontsize{8.5}{10.2}\selectfont,columns=fullflexible,keepspaces=true,showstringspaces=false,breaklines=true,aboveskip=0pt,belowskip=0pt]
{"subtasks": [{"id": "st_01", "name": "<capability-focused name>",
  "description": "<tested capability and meaningful distinction>",
  "answer_type": "choice", "modalities": ["text"],
  "sample_schema": {
    "input": {"fields": {"question": {"dtype": "text", "subtype": "question", "type": "str"}}},
    "output": {"fields": {"answer": {"dtype": "text", "subtype": "multiple_choice", "type": "str"}}}
  }, "keywords": ["<cue 1>", "<cue 2>", "<cue 3>"]}]}
\end{lstlisting}
\end{promptjson}

\begin{agentprompt}{PromptTeal}{Grounding Agent: Prompt Template}

You are the Grounding Agent for one evaluation subtask. Choose promising datasets and brief transformation routes for a Executor that specializes each sample. Judge source potential and feasibility, not final sample quality. Original content, annotations and correct QA are trusted facts; wrong options are not. Rich material or a few useful facts can support substantial reconstruction, new QA and evidence arrangements. Native task, modality or difficulty need not match when allowed transformations and supplied tool\_\allowbreak{}capabilities can bridge the gap. Do not weaken the target or reject a whole source for large edits or a few weak samples.
\par
\prompthead{Dataset decisions:}
Start with one Search and develop its pool. Batch independent Assess questions for sources needing inspection. Use the findings to choose promising sources for batch Plan. If information is already sufficient, direct Plan is valid.
\par
Plan success sets planned. Judge the returned blueprint's target fit, material basis, development value and dependencies. Explicitly accept a usable route with \textbf{manage\_\allowbreak{}candidates}(action=\textquotedbl{}ready\textquotedbl{}), investigate a concrete doubt, or revise a planned route with focused guidance. Ready is per dataset and can happen while others still need work.
\par
When all retained sources have usable accepted blueprints, all other sources are explicitly discarded, and no investigation/planning gap remains, call \textbf{finish\_\allowbreak{}grounding} with status=\textquotedbl{}ready\textquotedbl{}, selected dataset\_\allowbreak{}ids and a concise reason.
\par
\prompthead{Decision output and history:}
Return one native \textbf{submit\_\allowbreak{}grounding\_\allowbreak{}decision} call with history and actions. Each action has name and arguments matching its schema. Merge the prior history in working\_\allowbreak{}note with confirmed results from last\_\allowbreak{}round\_\allowbreak{}feedback and current state. Return the full updated history by dataset ID: preserve important earlier findings, record new results and remove discarded sources.
\par
Do not Assess, Plan or reserve a ready source again. Keep at most five datasets with current planned/ready blueprints, including reserved blueprints. Same-source revisions count once, and five is a ceiling, not a quota. Search again only after explicitly discarding the current pool and identifying a real material gap that justifies changed retrieval guidance. If no useful route can be established within budget, finish incomplete with the actual gap.
\par

\end{agentprompt}

\begin{promptjson}{PromptTeal}{Example: Grounding Decision}
\begin{lstlisting}[basicstyle=\rmfamily\fontsize{8.5}{10.2}\selectfont,columns=fullflexible,keepspaces=true,showstringspaces=false,breaklines=true,aboveskip=0pt,belowskip=0pt]
{"history":"","actions":[{"name":"search_datasets","arguments":{"guidance":""}}]}
\end{lstlisting}
\end{promptjson}

\begin{agentprompt}{PromptPurple}{Allocation Agent: Prompt Template}

You are the Allocation Agent in a benchmark construction system.
\par
Your responsibility is to allocate dataset samples to subtasks given the grounded instantiations from the Grounding Agent. If repeated reasonable attempts still fail, return the best effort result. In persistent failure cases, you may signal feedback for the Design Agent (e.g. quota or coverage constraints).
\par
\prompthead{Available tools:}
\begin{itemize}[leftmargin=1.15em,labelsep=0.4em,topsep=1.5pt,partopsep=0pt,itemsep=1pt,parsep=0pt]
\item \textbf{attempt\_\allowbreak{}allocation}()
\item \textbf{diagnose\_\allowbreak{}allocation}()
\item \textbf{case\_\allowbreak{}resolved}(decision\_\allowbreak{}rationale, summary)
\end{itemize}
\par
\prompthead{Configuration adjustment tools (use selectively):}
\begin{itemize}[leftmargin=1.15em,labelsep=0.4em,topsep=1.5pt,partopsep=0pt,itemsep=1pt,parsep=0pt]
\item \textbf{set\_\allowbreak{}subtask\_\allowbreak{}dataset\_\allowbreak{}allocation\_\allowbreak{}order}(subtask\_\allowbreak{}id, dataset\_\allowbreak{}order)
\item \textbf{set\_\allowbreak{}subtask\_\allowbreak{}allocation\_\allowbreak{}order}(order)
\item \textbf{refine\_\allowbreak{}quota}(quota\_\allowbreak{}for\_\allowbreak{}subtask)
\end{itemize}
\par
\prompthead{Operating loop:}
\begin{enumerate}[label=\arabic*),leftmargin=1.4em,labelsep=0.4em,topsep=1.5pt,itemsep=1pt,parsep=0pt]
\item Attempt allocation.
\item If allocation succeeds, stop (unfilled\_\allowbreak{}subtasks=0) and call \textbf{case\_\allowbreak{}resolved}(decision\_\allowbreak{}rationale, summary).
\item If allocation fails:
\begin{itemize}[leftmargin=1.15em,labelsep=0.4em,topsep=1.5pt,partopsep=0pt,itemsep=1pt,parsep=0pt]
\item Call \textbf{diagnose\_\allowbreak{}allocation} to understand the failure and obtain recommendations.
\item Apply a small number of targeted changes (order or quota).
\item Retry allocation.
\end{itemize}
\end{enumerate}
\par
Repeat a limited number of times. Avoid oscillating settings. Every tool call must include a brief decision\_\allowbreak{}rationale explaining the current state, the problem or opportunity, and why this is the best next action. Keep it concise. One to three sentences is enough.
\par
\tcblower
\prompthead{Current State (from ctx)}
\begin{itemize}[leftmargin=1.15em,labelsep=0.4em,topsep=1.5pt,partopsep=0pt,itemsep=1pt,parsep=0pt]
\item subtasks: \promptfield{\{subtasks\}}
\item datasets: \promptfield{\{datasets\}}
\item diagnosis and recommendations: \promptfield{\{last\_\allowbreak{}diagnosis\}}
\item allocation result: \promptfield{\{allocation\_\allowbreak{}result\}}
\item earlier run history: \promptfield{\{allocation\_\allowbreak{}agent\_\allowbreak{}history\_\allowbreak{}str\}}
\end{itemize}
\par

\end{agentprompt}

The resulting dataset--subtask assignments and counts are used to construct the Executor's sample indices.

\subsubsection{Executor Agents}

\label{supp:executor_agent_prompts}

We provide prompt excerpts for the Benchmark Executor's sample-level construction workflow (Section~\ref{sec:benchmark_realization}). \emph{Initial Review}, \emph{Repair}, and \emph{Final Review} together implement \emph{Quality Review}. Runtime requests supply the task contract, sample fields, media, and available tools; angle-bracketed strings denote illustrative JSON values.

Each stage returns one complete JSON object. Initial Review sends actionable revisions to Repair before Final Review; accepted drafts without changes go directly to Final Review. This draft-level repair is part of Quality Review; the model-feedback refinement rounds are specified in Appendix~\ref{supp:main_experiments}.

\begin{agentprompt}{PromptBlue}{Proposal: Prompt Template}

Explore ways to transform THIS original sample into a challenging test of the requested capability. Inspect its content, supplied media and correct QA/annotations to find usable details and rough ideas; then develop 2-3 substantively different directions. The original question, answer target and presentation may change; keep the exploration anchored in this sample.
\par
Explore different angles, connections or ways to use the material; explain the source opportunity, what could change and why that change could create a worthwhile, challenging test of the target. More facts, viewpoints or longer alternatives alone do not establish difficulty.
\par
Use original facts and ordinary stable knowledge to get ideas started. Search after an idea gives you a reason to supplement, check or enrich it with external knowledge or material. Prefer authoritative primary sources. Stop when further retrieval adds little; no search is needed when existing material suffices. Keep useful retrieved facts and actual URLs in evidence for internal verification, not mandatory display in final QA.
\par
proposals is an array of 2-3 objects with unique ids; each cites at least one existing original path.

\end{agentprompt}

\begin{promptjson}{PromptBlue}{Proposal: Response Example}

\begin{lstlisting}[basicstyle=\rmfamily\fontsize{8.5}{10.2}\selectfont,columns=fullflexible,keepspaces=true,showstringspaces=false,breaklines=true,aboveskip=0pt,belowskip=0pt]
{"action": "finish", "proposals": [
  {"id": "p1", "goal": "<target-aligned opportunity>",
   "source_refs": ["original.input.<field>"],
   "strategy": "<source inspiration, possible change and difficulty opportunity>"},
  {"id": "p2", "goal": "<different opportunity>",
   "source_refs": ["<existing original path>"],
   "strategy": "<another possible direction>"}
], "evidence": []}
\end{lstlisting}

\end{promptjson}


\begin{agentprompt}{PromptBlue}{Design: Prompt Template}

Develop ONE concrete plan to transform the original sample into a challenging target-aligned evaluation item. Inspect the source and media, use the Proposal brainstorm and supplementary facts, and choose a promising point of departure in this sample. Work with available facts and original media; this role does not search.
\par
Plan material and QA together: what to retain, restructure or supplement; what the solver receives; what they must do; and what supports the answer. Keep tested knowledge and conclusions for the solver to supply, while providing necessary situation-specific premises. Explain the arrangement that makes the target necessary and challenging.
\par
For reasoning, identify the relationship, interacting condition or evidence combination that determines the conclusion. For choices, outline the key distinction and plausible confusions among close alternatives, with a clear basis for excluding them. Keep rivals credible in the relevant context; leave concrete options to Generation.
\par
Preserve useful complexity and an adjudicable gold, avoiding accidental ambiguity, missing required premises or verbose QA. Resolve the central design decisions while leaving exact wording, complete choices and the final answer value or label to Generation.

\end{agentprompt}

\begin{promptjson}{PromptBlue}{Design: Response Example}

\begin{lstlisting}[basicstyle=\rmfamily\fontsize{8.5}{10.2}\selectfont,columns=fullflexible,keepspaces=true,showstringspaces=false,breaklines=true,aboveskip=0pt,belowskip=0pt]
{"decision": "proceed", "design": {
  "goal": "<target capability and intended challenge>",
  "source_refs": ["<existing original/evidence dot path>"],
  "strategy": "<concrete whole-sample transformation plan>"}}
\end{lstlisting}

\end{promptjson}


\begin{agentprompt}{PromptTeal}{Generation: Prompt Template}

Transform the original sample according to the supplied design and return ONE complete evaluation item. Realize its material changes, target-capability test and difficulty mechanism; complete the material, question, applicable alternatives and answer in the required schema. Use judgment on implementation details left open by the plan, without selecting a new route.
\par
Write a natural, clear evaluation item from the design. Let the evaluation target determine which conditions to provide and which knowledge, judgment or inference the solver must supply. Keep design rationale and author-side checks internal.
\par
Apply the sample's format, answer-type, modality and natural-expression requirements. Before returning, check that execution preserved the planned challenge, introduced no answer leakage, and yields a justified gold. Determine the answer and brief internal support from the finished item, not the old label.

\end{agentprompt}

\begin{promptjson}{PromptTeal}{Generation: Response Example}

\begin{lstlisting}[basicstyle=\rmfamily\fontsize{8.5}{10.2}\selectfont,columns=fullflexible,keepspaces=true,showstringspaces=false,breaklines=true,aboveskip=0pt,belowskip=0pt]
{"status": "ready",
 "candidate": {"input": {"question": "<str>"},
               "output": {"answer": "<str>"}},
 "support": {"answer_refs": ["<existing original/evidence path>"],
             "answer_justification": "<brief evidence-to-answer basis>"}}
\end{lstlisting}

\end{promptjson}


\begin{agentprompt}{PromptTeal}{Replan: Prompt Template (When Generation Is Blocked)}

Recover from the supplied construction blocker before initial review. Read the existing design and factual inputs; return a complete supported design resolving that blocker within the same evaluation target and hard constraints.
\par
Base the revised transformation on original material, correct QA, reliable knowledge and existing evidence. No new searches; prefer a viable source-grounded route when an external premise is unavailable. Return the replan template with evidence\_\allowbreak{}requests=[] inside design.

\end{agentprompt}


\begin{agentprompt}{PromptPurple}{Initial Review: Prompt Template}

Review the transformed item against task\_\allowbreak{}contract and benchmark goal using concrete evidence. Check target alignment, gold, answerability, challenge, leakage and sample rules. Assess solving from final input/media and task-appropriate knowledge. Design is intent, not proof.
\par
Initial review (review\_\allowbreak{}mode=refinement\_\allowbreak{}strategy): use the concrete draft to identify shortcomings and errors, then give Repair a specific improvement plan. Inspect the actual solving route: does it exercise the target with meaningful challenge, or merely copy an explanation or reject conspicuous errors?
\par
For each important problem, locate it, explain its cause and prescribe how to fix THIS candidate. Specify the affected material, question focus, alternatives or answer and the intended result; do not stop at 'increase difficulty' or 'integrate evidence'. Give an observable acceptance check that tests the defect, not just completion of an edit.
\par
If the sample already meets the target with meaningful challenge, supported gold and no worthwhile correction, return accept with changes:[]. Reject/inconclusive only for a concrete hard obstacle with no viable grounded correction.

\end{agentprompt}

\begin{promptjson}{PromptPurple}{Initial Review: Response Example}

\begin{lstlisting}[basicstyle=\rmfamily\fontsize{8.5}{10.2}\selectfont,columns=fullflexible,keepspaces=true,showstringspaces=false,breaklines=true,aboveskip=0pt,belowskip=0pt]
{"decision": "revise", "reason": "<concrete diagnosis>",
 "changes": [{"path": "candidate",
              "problem": "<observed defect and cause>",
              "change": "<specific executable correction>",
              "check": "<observable success condition>"}]}
\end{lstlisting}

\end{promptjson}


\begin{agentprompt}{PromptTeal}{Repair: Prompt Template}

Improve the existing transformed sample using the initial review's concrete plan in ONE revision. Preserve what works; the changes may revise the whole context, question, alternatives and gold together. The review may correct a weak original design; follow that justified adjustment within the same evaluation target and available material.
\par
Meet the purpose and acceptance check of each change. Removing a sentence is not sufficient when the same shortcut survives elsewhere; a revised option set must not recreate the diagnosed problem with different isolated errors. If only format is defective, preserve the valid content while fixing its structure.
\par
Recompute gold (using current options for choices), brief support and applicable media outputs, and apply the same schema, answer-type, modality and concise-QA requirements as Generation. Do not search again. Return the complete result in the Generation output format, not a patch; blocked is only for an unavailable essential premise/media or incompatible hard constraint.

\end{agentprompt}


\begin{agentprompt}{PromptPurple}{Final Review: Prompt Template}

Final review (review\_\allowbreak{}mode=final\_\allowbreak{}validation): judge the current candidate. For choices, read gold from candidate.output.answer and verify it against current options.
\par
Require supported, adjudicable gold and a meaningful test of the target. Reject incorrect/unsupported/ambiguous gold, missing essential premises, essential task/modality mismatch, hard-limit violations or a target-bypassing shortcut. Direct lookup is valid when requested.
\par
Difficulty matters: trivial restatement or a token use of the target does not meet a challenging benchmark's goal. Ground that judgment in the actual task and solving route, without invented scores. No new design, search or major rewrite.
\par
Allow at most 4 certain exact minor fixes, before/after \textless{}=80 characters and before occurring once: format,typo,factual\_\allowbreak{}wording,reference\_\allowbreak{}format. Use permitted input paths or support.answer\_\allowbreak{}refs.\textless{}index\textgreater{}; factual\_\allowbreak{}wording requires existing evidence\_\allowbreak{}refs. No gold/media/whole-passage/evidence-dependency changes. Reject major defects; safe fixes get local checks, not another review.

\end{agentprompt}

\begin{promptjson}{PromptPurple}{Final Review: Response Example}

\begin{lstlisting}[basicstyle=\rmfamily\fontsize{8.5}{10.2}\selectfont,columns=fullflexible,keepspaces=true,showstringspaces=false,breaklines=true,aboveskip=0pt,belowskip=0pt]
{"decision": "revise", "reason": "<decisive evidence or release blocker>",
 "minor_fixes": [{"kind": "typo", "path": "<permitted text path>",
                  "before": "<exact old text>", "after": "<replacement>"}]}
\end{lstlisting}

\end{promptjson}


\subsubsection{Refinement}
\label{supp:refinement_agent_prompts}

We provide condensed prompts for the Guidance Agent used in Refinement with Feedback. Its four stages select correct controls, review evaluated cases, develop guidance from verified errors, and review the resulting guidance. Shared instructions apply throughout. The final guidance is passed to the Executor for construction from original source samples.

\begin{agentprompt}{PromptBlue}{Shared Instructions}

You are the Guidance Agent for Refinement with Feedback. You analyze evaluation evidence to derive subtask-specific directions for constructing harder, valid questions from raw original dataset samples. The Executor receives the raw sample and its media, not the old evaluated question or this analysis.
\par
Base all guidance on verified wrong answers. Use correct answers as controls. Preserve the user's capability, uniquely defensible reference answers, and usable media. Separate observed facts, supported interpretations, and untested hypotheses; a selected answer does not reveal private model reasoning.
\par
If no supported wrong observation remains, produce no guidance.

\end{agentprompt}

\begin{agentprompt}{PromptTeal}{Exploration: Select Correct Controls}

Use the supplied evaluation statistics and case catalog to choose correct controls for the selected wrong cases. Prefer closely matched material, targets, and scopes that distinguish competing explanations. Choose controls to investigate the observed errors.
\par
Catalog descriptions and filenames are selection aids, not proof of media content. Keep whole-cohort statistics separate from findings about inspected cases, and do not claim to have reviewed omitted cases.

\end{agentprompt}

\begin{agentprompt}{PromptBlue}{Case Review: Check Evidence and Answers}

Inspect the evaluated question and supplied media. Only the evaluated input is solver-visible; source answers and generation notes are author context. Verify the reference against the evidence available for the question. Mark it questionable or unverifiable when a reliable judgment is not supported; missing essential media make it unverifiable.
\par
For each verified wrong answer, identify the erroneous or unsupported claim in the selected option, the cue that makes it plausible, and the evidence or missing relation that distinguishes it from the reference. Describe the raw-source conditions under which this distinction could recur. Treat explanations of the model's error as hypotheses and consider alternatives.
\par
For correct controls, record the distinctions already resolved and possible shortcuts. Derive no hardening direction from a correct answer or an unsupported reference. When a supplied transcript is designated as authoritative, use its utterances and speaker labels as evidence.

\end{agentprompt}

\begin{agentprompt}{PromptPurple}{Synthesis: Develop Construction Guidance}

Synthesize guidance from the reviewed wrong cases, using correct controls to test and bound the interpretation. Group errors by the failed evidence relationship and what made a distractor plausible, rather than by topic alone.
\par
Each direction must identify the tempting cue, decisive distinction, and observable conditions in the raw material. Use independent, reference-supported errors to justify a recurring pattern; keep limited evidence as a narrow, conditional hypothesis.
\par
Write concise English guidance within the supplied length limit. The Executor must be able to apply it using only a raw source sample and the subtask. Keep supporting case IDs in the analysis fields, outside the injected guidance. Preserve the target capability and answer validity.

\end{agentprompt}

\begin{agentprompt}{PromptTeal}{Guidance Review: Check and Revise Guidance}

Review the analysis and guidance against the case reviews and evaluation statistics. Remove directions without verified wrong-answer support, overgeneralizations, and explanations contradicted by correct controls. Exclude mechanisms based on questionable or unverifiable references.
\par
For every direction, construct the easiest question that would satisfy its wording. If that question avoids the observed difficulty, tighten the applicability and exclude the easy variant, or remove the direction. Ensure that the revised advice transfers to other raw samples while preserving answer validity.
\par
Return the corrected synthesis JSON, with \textbf{review\_\allowbreak{}notes} recording substantive revisions or remaining limitations. The guidance remains unvalidated until tested in a later construction and evaluation round.

\end{agentprompt}

\begin{promptjson}{PromptTeal}{Reviewed Guidance: Response Example}

\begin{lstlisting}[basicstyle=\rmfamily\fontsize{8.5}{10.2}\selectfont,columns=fullflexible,keepspaces=true,showstringspaces=false,breaklines=true,aboveskip=0pt,belowskip=0pt]
{"readiness": "exploratory",
 "mechanisms": [{
   "id": "m1", "status": "candidate_hypothesis",
   "finding": "<supported evidence relationship>",
   "support_case_ids": ["<verified wrong case ID>"],
   "control_case_ids": ["<matched correct control ID>"],
   "applicability": "<observable conditions in raw material>",
   "weakening_rewrite": "<closest easy variant to exclude>",
   "hardening_direction": "<transferable construction direction>"
 }],
 "unsupported_directions": [],
 "guidance": "<self-contained English guidance for new items>",
 "validation_plan": ["<validity and full-set performance checks>"],
 "review_notes": ["<substantive revisions or limitations>"]}
\end{lstlisting}

\end{promptjson}

\noindent If no supported wrong-answer evidence remains after review, \textbf{readiness} is \textbf{insufficient\_\allowbreak{}evidence}, \textbf{mechanisms} is empty, and \textbf{guidance} is an empty string.

\section{Experimental Settings}
\label{supp:experimental_settings}

\subsection{Main Experiments}
\label{supp:main_experiments}
\noindent\textbf{Benchmark construction.}
We construct 16 benchmarks with a target of 200 items per benchmark, using General-Bench as the default source pool. GPT-6.1-Sol is the backbone for the Benchmark Planner, Benchmark Executor, and Guidance Agent in the main experiments.

\noindent\textbf{Refinement feedback.}
At the start of each refinement round, we evaluate the current reviewed benchmark using Qwen3.8-27B for Image--Text requirements and Qwen3.8-Omni-Flash for Audio--Text and Image--Audio--Text requirements. Their responses provide the feedback for refinement. For audio items, the Guidance Agent reviews transcripts, while the feedback models receive audio inputs.

\noindent\textbf{Number of refinement rounds.}
The round count excludes initial construction and Quality Review: zero rounds corresponds to the initial benchmark $\mathcal{E}^{(0)}$. For the benchmark versions reported in Table~\ref{table:benchmark_quality}, Target-Speaker Tracking and Lyric Interference use zero refinement rounds. Narrative Causality, Geographic Reasoning, Animal Reasoning, Audio Misdirection, and Visual Detail QA use two rounds. The remaining nine benchmarks use one round.

\noindent\textbf{Guidance and reconstruction.}
For each subtask, the Guidance Agent checks the reference answers and evidence of incorrectly answered items and compares verified errors with closely matched correct responses. It summarizes supported findings as subtask-specific construction guidance and reviews that guidance before passing it to the Executor. The Executor then constructs new items from the original source pool using the guidance, while retaining the benchmark requirement, target size, and Quality Review procedure. If no supported wrong-answer evidence remains, the Guidance Agent produces no guidance for that subtask. The condensed prompts are provided in Appendix~\ref{supp:refinement_agent_prompts}.

\subsection{Ablation Studies}
\label{supp:ablation_studies}

Both the backbone comparison and the component ablation use the same three benchmarks: Artwork Reasoning (Image--Text), Audio Answerability (Audio--Text), and Scene Misdirection (Image--Audio--Text), with a target of 200 items per benchmark. We assess benchmark quality using Format Quality, Evidence Correspondence, Task Alignment, and Answer Validity, following the LLM-as-Judge protocol in Appendix~\ref{sec:llm as judge}.

\noindent\textbf{Refinement feedback.}
In runs with refinement, the Guidance Agent uses the backbone of the corresponding construction run. Feedback models follow the modality-specific setting in Appendix~\ref{supp:main_experiments}.

\subsubsection{Direct Generation}
\label{supp:direct_generation}

Direct Generation uses a single LLM-based agent for dataset retrieval, item construction, and self-review. We run it separately with GPT-6.1-Sol, Gemini-3.8-Flash, DeepSeek-V4.1-Flash, and Qwen3.5-397B-A17B, matching the backbones in Table~\ref{table:generation_backbone}. Direct Generation and \methodname{} receive the same benchmark requirements and share the same transformation tool library.

\noindent\textbf{Dataset selection.}
Given the complete benchmark requirement, the agent writes Python code to search a local catalog of dataset cards using keywords it chooses. It may filter cards by modality or rank them by keyword matches. After reviewing shortlisted cards and optionally inspecting source examples, it selects one to three datasets for the entire requirement and provides a rationale for its choices.

\noindent\textbf{Item construction and evaluation.}
The system samples source examples from the selected datasets in round-robin order. For each source example, the agent constructs a four-option multiple-choice item and its reference answer. It may invoke tools to gather evidence, check the draft, and revise it over multiple interactions before submission. In the DeepSeek, Gemini, and Qwen runs, optional web searches during this stage are handled by GPT-6.1-Sol. Submitted items must pass format and provenance checks before being used for model evaluation against their reference answers.

\subsubsection{Component Ablation}
\label{supp:component_ablation}

The full \methodname{} setting uses the Design, Grounding, and Allocation agents in the Benchmark Planner, followed by item construction and Quality Review in the Benchmark Executor. In this study, the full setting also applies one round of Refinement with Feedback after evaluating the initial benchmark. We fix the backbone to GPT-6.1-Sol and remove or replace one component at a time, as in Table~\ref{tab:component_ablation}.

\noindent\textbf{w/o Design.}
We bypass the Design Agent and represent the original requirement as a single subtask with the required input modalities and question--answer format. The Grounding Agent, Allocation Agent, and subsequent execution stages operate on that subtask.

\noindent\textbf{w/o Grounding.}
For each subtask produced by the Design Agent, we invoke the full system's candidate retriever using the subtask text as the query. The retriever combines BM25 and TF--IDF cosine similarity over two channels: native task descriptions and metadata describing available source material. This variant omits dataset preference generation, source sample assessment, and dataset-level blueprint development, corresponding to \emph{Preference}, \emph{Assess}, and \emph{Plan} in the full Grounding workflow. Up to three top-ranked datasets and their retrieval scores are passed directly to the Allocation Agent.

\noindent\textbf{w/o Executor.}
We replace the structured Benchmark Executor construction workflow with a lightweight tool-using GPT-6.1-Sol agent. The agent receives each allocated source example and its dataset-level blueprint, chooses tools from the shared library, and constructs and checks the item. The resulting submission still undergoes the original pipeline's independent Final Review.

\noindent\textbf{w/o Refinement.}
The full setting performs one feedback-based refinement round after evaluating the initial benchmark. This variant uses zero refinement rounds and evaluates the initially reviewed benchmark $\mathcal{E}^{(0)}$ directly. Initial item construction and Quality Review, including draft repair when needed, are retained.

\noindent\textbf{Final Review pass rate.}
All component variants retain the shared Final Review stage. The pass rate is the percentage of submitted candidates accepted by this stage, matching the definition in Table~\ref{tab:component_ablation}. This construction-stage acceptance measure is reported separately from the four LLM judge scores.


\section{Evaluation Protocols}

\subsection{Human Evaluation}
\label{supp:human_evaluation}

For each benchmark, two reviewers independently assess the same 50 items along two dimensions, assigning \textbf{Pass} or \textbf{Fail} to each.
An item is marked \textbf{Fail} on a dimension if either reviewer assigns \textbf{Fail}.

\noindent\textbf{Evidence Correspondence.}
An item receives \textbf{Pass} if specific information in the provided text, images, or audio contributes to interpreting the question, reasoning through it, or determining the answer. It receives \textbf{Fail} if the materials are irrelevant or serve only as background or decoration.

\noindent\textbf{Task Alignment.}
An item receives \textbf{Pass} if at least one step in a reasonable solution requires the capability specified in the original benchmark requirement. It receives \textbf{Fail} if the item can reasonably be solved without exercising that capability.

\noindent\textbf{Item Correctness.}
Separately, human reviewers verify the reference answers for all generated items with GPT-6.1-Sol assistance and make the final correctness judgment.
An item is counted as correct when the question is answerable as written and its reference answer satisfies the stated criterion, given the supplied material and task-appropriate knowledge.
For multiple-choice items, the keyed option must be correct and no other option may also satisfy that criterion.
Wrong keys, unanswerable questions, and competing correct options are counted as correctness errors.

\subsection{LLM-as-Judge}

\label{sec:llm as judge}

The evaluation protocol covers all \textbf{3,200 generated benchmark items}, with \textbf{200 items for each of 16 benchmark requirements}. Two judges, GPT-6.1-Sol and Seed-2.1-Turbo, independently assess each item using the same evidence. For each dimension they can assess, the judges assign an integer score from 1 to 5 and provide a brief justification in English.
For audio items, the judges review transcripts and the supplied audio production information. Benchmark solvers receive the audio. Both judges use the same four scoring rubrics. Each benchmark requirement also has its own guidance, developed from a shared template and rules specific to that requirement.

\noindent\textbf{Score Aggregation.}
Each judge first assesses all 200 items for a requirement individually. For each dimension, we then calculate each judge's mean over their numeric item scores. The \textbf{final score is the equal-weight average of these two means}, with each judge contributing 50\%. We multiply the result by 20 to report it on a 100-point scale.

\noindent\textbf{Scoring Rubrics.}
The descriptions are condensed. The five scoring anchors for each dimension retain the original prompt wording.

\begin{ratingbox}{Format Quality}

Check whether the required fields, instructions, options, and answer format are clear, consistent, and usable.

\begin{enumerate}[label=\textbf{\arabic*:},leftmargin=1.5em,labelsep=0.45em,topsep=3pt,partopsep=0pt,itemsep=2pt,parsep=0pt]

\item Broken or unreadable; essential question/answer/required options are missing.

\item Major formatting or expression defects substantially obstruct reliable use.

\item Usable with noticeable inconsistencies or unclear expression.

\item Clear and usable with only minor presentation problems.

\item Clear, complete, consistent, and ready to present and score.

\end{enumerate}

\end{ratingbox}

\begin{ratingbox}{Evidence Correspondence}

Assess how the question uses specific content from the supplied material in its setup, inquiry, or solution.

\begin{enumerate}[label=\textbf{\arabic*:},leftmargin=1.5em,labelsep=0.45em,topsep=3pt,partopsep=0pt,itemsep=2pt,parsep=0pt]

\item The question does not use the supplied material at all.

\item Only a superficial thematic connection; material is mostly decorative.

\item Some material content is used, but the connection is loose or limited.

\item The question clearly uses information provided by the material.

\item The question naturally and specifically incorporates the material's content.

\end{enumerate}

\end{ratingbox}

\begin{ratingbox}{Task Alignment}

Assess whether the operations involved in solving the item meaningfully exercise the capability specified in the original benchmark requirement.

\begin{enumerate}[label=\textbf{\arabic*:},leftmargin=1.5em,labelsep=0.45em,topsep=3pt,partopsep=0pt,itemsep=2pt,parsep=0pt]

\item Does not test the intended capability, or tests an unrelated capability.

\item Mainly tests a different capability; the target capability is only incidental.

\item Partly tests the target capability, with a substantial actual omission or mismatch.

\item Tests the target capability well, with a minor, concrete capability mismatch.

\item Clearly and meaningfully tests the target capability; ordinary clues or ease of solving do not preclude full marks.

\end{enumerate}

\end{ratingbox}

\begin{ratingbox}{Answer Validity}

Check whether the question is solvable as written, the reference answer satisfies the stated criterion, and any competing answer also meets that criterion. Record a missing score when an unresolved fact prevents a reliable judgment.

\begin{enumerate}[label=\textbf{\arabic*:},leftmargin=1.5em,labelsep=0.45em,topsep=3pt,partopsep=0pt,itemsep=2pt,parsep=0pt]

\item A verified wrong key, or a fundamentally unsolvable item as written.

\item Major ambiguity or another major answer defect makes the single-label key unreliable, normally two equally correct options or an answer-changing omission.

\item The same core answer is uniquely correct and reliably scorable, but a clear local wording, precision or nonessential-detail defect requires correction before publication. Name the concrete repair and why it cannot change the key. Several interacting nondecisive imprecisions can qualify. This is not a compromise for uncertainty, a false required clause or a rival correct answer.

\item The same core answer is uniquely correct. One slight wording/precision flaw does not require repair for reliable use. Identify it; otherwise give 5.

\item The answer is correct, clear and reliably scorable; no actual validity flaw.

\end{enumerate}

\end{ratingbox}


\section{Limitations}
\label{supp:limitations}
Despite the promising results, our current framework still has several minor limitations. 
First, the constructed benchmarks are mainly validated on representative tasks, while more domain-specific scenarios can be further explored. In future work, we will expand the benchmark pool to cover broader domains and more diverse user requirements. 
Second, \methodname{} is partly bounded by the coverage of available tools. Since our system is built in a modular agentic manner, it can be naturally extended by seamlessly integrating more tools into the tool library. In future work, we will expand the available tool set and enable the agent to flexibly select, combine, and invoke different tools.
Third, although \methodname{} can substantially reduce human effort, light-weight human review is still useful. We plan to incorporate more interactive feedback mechanisms so that experts can efficiently refine constructed benchmarks with minimal additional cost. Overall, these limitations do not affect the main conclusion of this work, but point to practical directions for making autonomous benchmark construction more general, flexible, and user-friendly.

\par\medskip
\section{Benchmark Visualization}
\label{supp:benchmark_visualization}
\begingroup
\raggedbottom
\definecolor{CaseInk}{HTML}{293B49}
\definecolor{CaseMuted}{HTML}{667783}
\definecolor{CaseImage}{HTML}{67517D}
\definecolor{CaseAudio}{HTML}{355E78}
\definecolor{CaseOmni}{HTML}{8A6138}
\definecolor{CaseWrong}{HTML}{A93E40}
\definecolor{CaseRight}{HTML}{2D6D58}
\newcommand{\casefield}[1]{\par\addvspace{3pt}\noindent\textbf{\color{CaseInk}#1}\par\nobreak}
\newcommand{\caseoption}[2]{%
  \par\noindent{\hangindent=1.6em\hangafter=1%
  \makebox[1.6em][l]{\textbf{#1.}}#2\par}%
}
\newcommand{\caseanswer}[3]{%
  \par\nobreak\vspace{4pt}\nobreak
  {\color{CaseMuted!35}\hrule height 0.35pt}\nobreak
  \vspace{3pt}\nobreak\noindent
  \textbf{#1}\par\nobreak
  \noindent\textcolor{CaseWrong}{\textbf{Prediction: #2}}\hfill
  \textcolor{CaseRight}{\textbf{Reference: #3}}\par
}
\newcommand{\casegroup}[2]{%
  \par
  \ifdim\dimexpr\pagegoal-\pagetotal\relax<210pt\newpage\fi
  \subsection{#1}
  \noindent #2\par\nobreak
}
\newtcolorbox{benchmarkcase}[4][]{
  enhanced,breakable,arc=1mm,boxrule=0.5pt,
  colback=white,colframe=#2!45!white,
  colbacktitle=#2!10!white,coltitle=CaseInk,
  colbacklower=#2!3!white,
  borderline west={1.7pt}{0pt}{#2},
  title={#3\quad #4},
  title after break={#3\quad #4\space(continued)},
  fonttitle=\rmfamily\bfseries\normalsize,
  fontupper=\rmfamily\fontsize{9.5}{11}\selectfont,
  fontlower=\rmfamily\fontsize{9.5}{11}\selectfont,
  colupper=CaseInk,collower=CaseInk,
  left=8pt,right=8pt,top=4pt,bottom=4pt,boxsep=1.5pt,
  before skip=6pt,after skip=6pt,lines before break=5,
  before upper={\setlength{\parindent}{0pt}\setlength{\parskip}{1.5pt}\raggedright},
  before lower={\setlength{\parindent}{0pt}\setlength{\parskip}{1.5pt}\raggedright},
  breakable=true,
  #1
}
\noindent We show one failure case for each of the 16 benchmarks, with verbatim
text and original image order. Transcript ellipses mark omitted speech.

\casegroup{Image--Text Cases}{For each of the seven Image--Text benchmarks, we present one example that Qwen3.8-Flash answered incorrectly. Each example includes the original image(s), question, answer options, and textual context where applicable. The bottom row shows the model prediction in red and the reference answer in green.}

\begin{benchmarkcase}{CaseImage}{01}{Artwork Reasoning}
\noindent\begin{minipage}{\linewidth}
\centering
\includegraphics[width=\linewidth,height=2.3in,keepaspectratio]{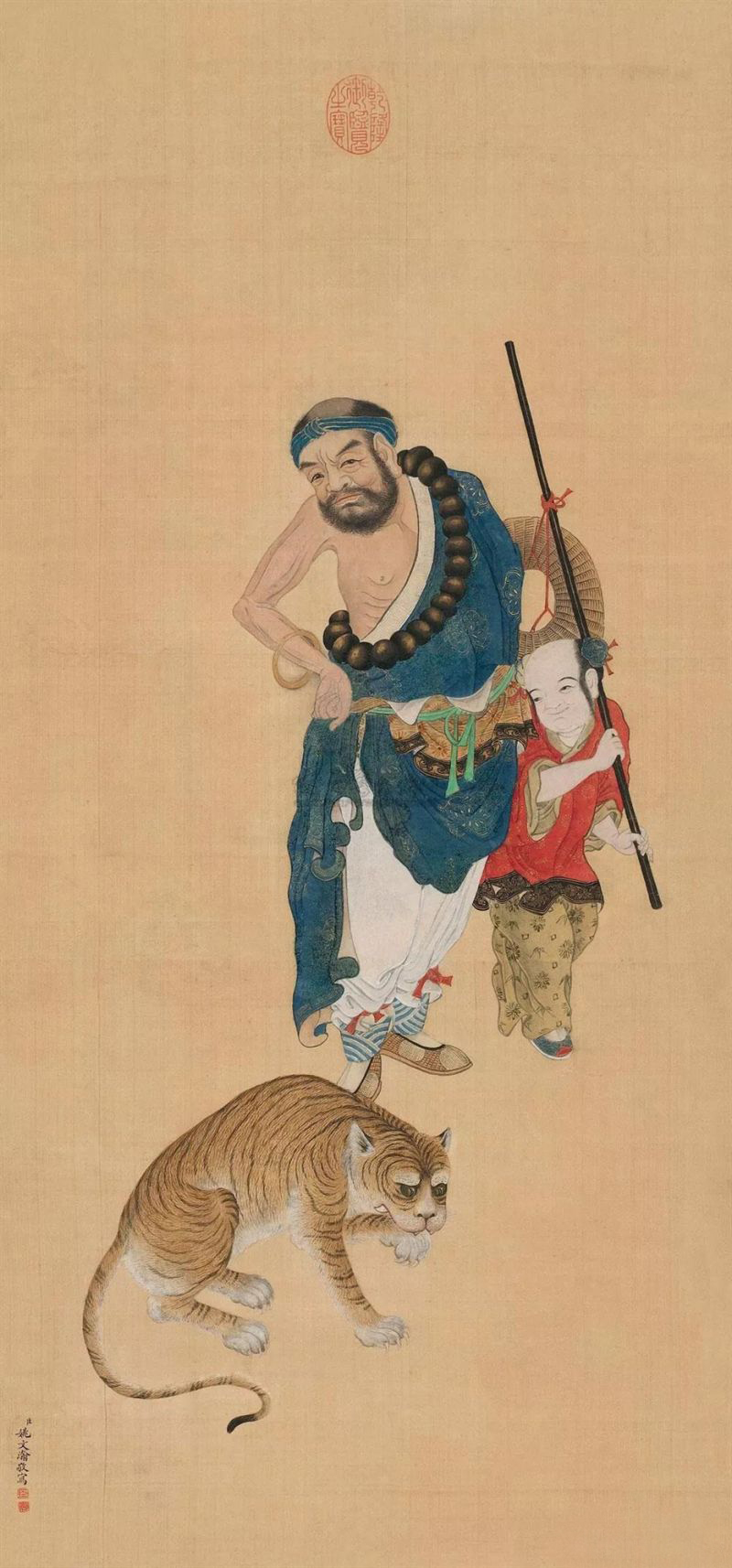}
\end{minipage}\par\smallskip
\casefield{Question}
The painter of the pictured work also made a painting depicting drink vendors. Which account best explains that painting's relationship to an earlier pictorial tradition and the historical significance of its accessories?
\caseoption{A}{It adapts Mingyuan Dushi, attributed to Liu Songnian, preserving Song-period ceramic glazes and dress as an antiquarian reconstruction rather than updating them for court life.}
\caseoption{B}{It adapts Qingming Shanghe Tu, attributed to Zhang Zeduan, preserving Song-period ceramic glazes and dress as an antiquarian reconstruction rather than updating them for court life.}
\caseoption{C}{It adapts Mingyuan Dushi, attributed to Liu Songnian, introducing Qing ceramic glazes and court fashions rather than preserving the material culture of the earlier model.}
\noindent\begin{minipage}{\linewidth}
\caseoption{D}{It adapts Qingming Shanghe Tu, attributed to Zhang Zeduan, introducing Qing ceramic glazes and court fashions rather than preserving the material culture of the earlier model.}
\caseanswer{Qwen3.8-Flash}{A}{C}
\end{minipage}\par

\end{benchmarkcase}

\begin{benchmarkcase}{CaseImage}{02}{Textual Misdirection}
\noindent\begin{minipage}{\linewidth}
\centering
\includegraphics[width=\linewidth,height=2.1in,keepaspectratio]{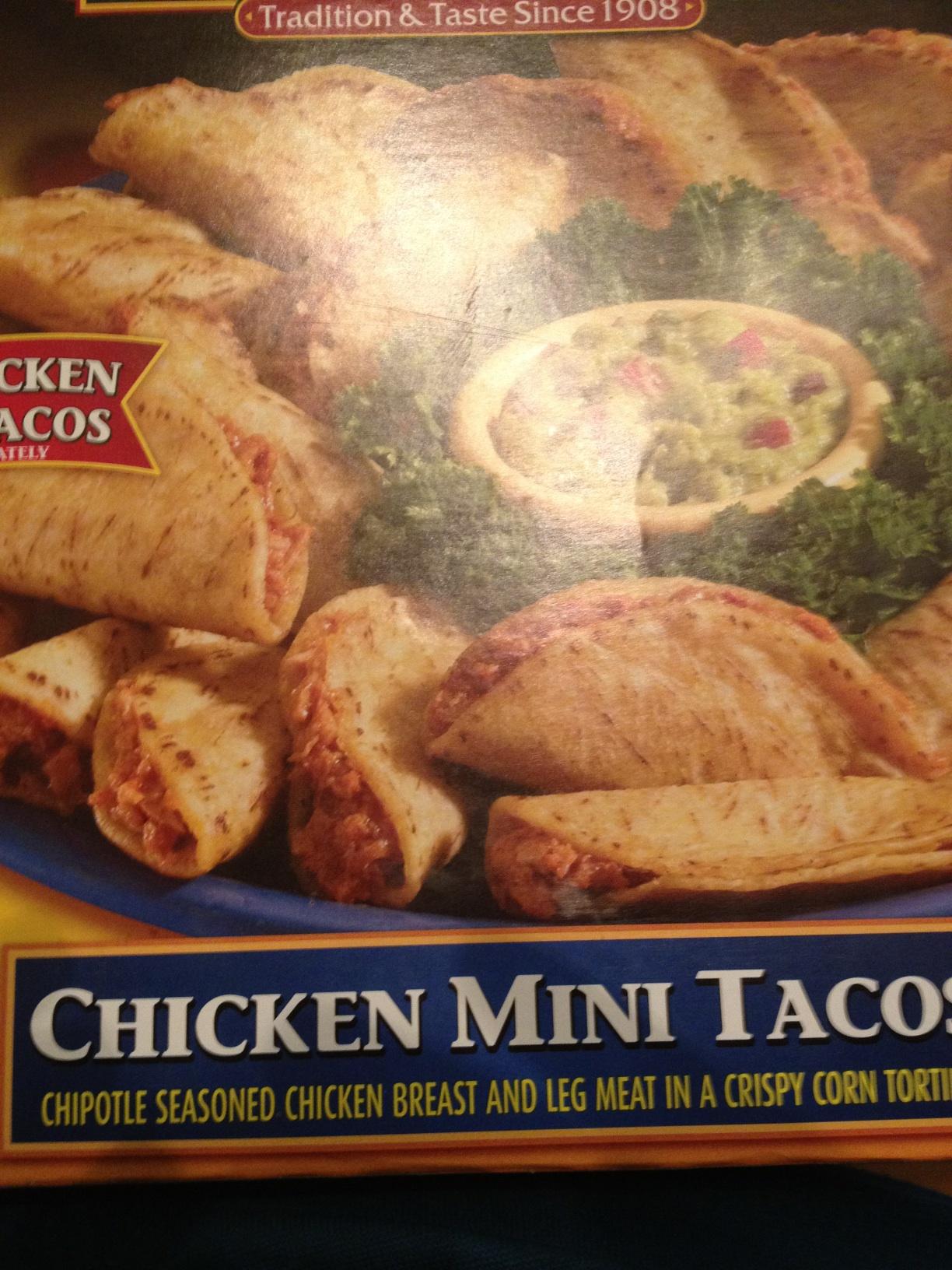}
\end{minipage}\par\smallskip
\casefield{Context}
The tacos are clearly package artwork, but the dip bowl has a shaded rim and the surrounding leaves have a crinkled texture. Those realistic details suggest that an actual bowl of dip and leafy garnish were placed on top of the box for the photograph.

\casefield{Question}
Which interpretation of the dip bowl and surrounding leafy greens is best supported by the photograph?
\caseoption{A}{The bowl and greens are both physical objects resting on top of the printed taco image.}
\caseoption{B}{The bowl and greens are both part of the printed serving illustration on the package.}
\caseoption{C}{The greens are part of the printed serving illustration, but the bowl is a physical object resting on the package.}
\noindent\begin{minipage}{\linewidth}
\caseoption{D}{The bowl is part of the printed serving illustration, but the greens are physical garnish resting on the package.}
\caseanswer{Qwen3.8-Flash}{A}{B}
\end{minipage}\par

\end{benchmarkcase}

\begin{benchmarkcase}{CaseImage}{03}{Visual Answerability}
\noindent\begin{minipage}{\linewidth}
\centering
\includegraphics[width=\linewidth,height=2.4in,keepaspectratio]{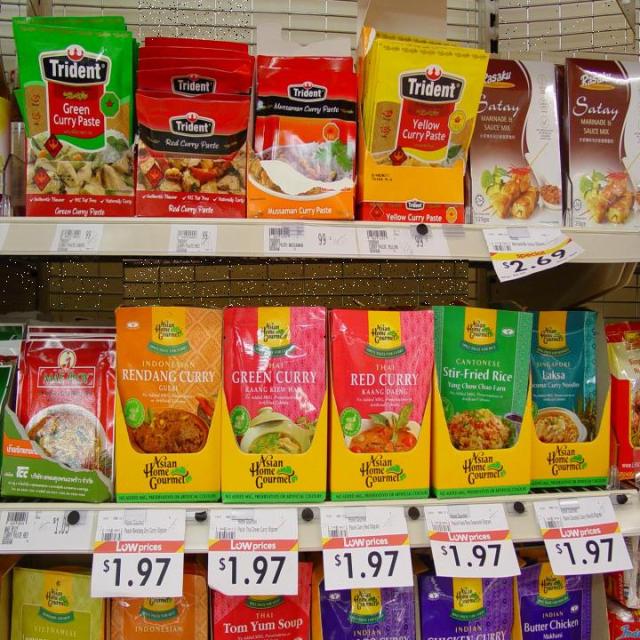}
\end{minipage}\par\smallskip
\casefield{Question}
Consider the orange Rendang Curry carton immediately left of the Green Curry carton on the middle shelf. Is at least one additional retail seasoning package directly behind that orange carton, within its left-to-right span and between its supporting shelf and the shelf immediately above? Exclude the orange carton itself and packages on other shelf levels.
\caseoption{A}{The image confirms that no additional package is present in that space.}
\caseoption{B}{The image does not establish whether any additional package is present in that space.}
\noindent\begin{minipage}{\linewidth}
\caseoption{C}{The image confirms that at least one additional package is present in that space.}
\caseanswer{Qwen3.8-Flash}{A}{B}
\end{minipage}\par

\end{benchmarkcase}

\begin{benchmarkcase}{CaseImage}{04}{Geographic Reasoning}
\noindent\begin{minipage}{\linewidth}
\centering
\includegraphics[width=\linewidth,height=1.7in,keepaspectratio]{figs/benchmark_examples/query12_gkv0194_1.png}
\end{minipage}\par\smallskip
\casefield{Question}
Using the map, select the northernmost of the four labeled states. Which reservoir\textendash{}conveyance\textendash{}receiver account correctly describes a diversion system in that state? Consider normal operation through the listed tunnel or pipeline system, including its powerhouse where applicable, not spillway or direct dam releases. The receiver must be the first receiving water at the system's discharge, not a later storage destination.
\caseoption{A}{Santeetlah Reservoir \textemdash{} Santeetlah pipeline/tunnel and powerhouse \textemdash{} Cheoah River}
\caseoption{B}{Santeetlah Reservoir \textemdash{} Santeetlah pipeline/tunnel and powerhouse \textemdash{} Cheoah Reservoir on the Little Tennessee River}
\caseoption{C}{Schoharie Reservoir \textemdash{} Shandaken Tunnel \textemdash{} Esopus Creek}
\noindent\begin{minipage}{\linewidth}
\caseoption{D}{Schoharie Reservoir \textemdash{} Shandaken Tunnel \textemdash{} Ashokan Reservoir}
\caseanswer{Qwen3.8-Flash}{D}{C}
\end{minipage}\par

\end{benchmarkcase}

\begin{benchmarkcase}[breakable=true]{CaseImage}{05}{Narrative Causality}
\noindent\begin{minipage}[t]{.485\linewidth}
\centering
\includegraphics[width=\linewidth,height=1.5in,keepaspectratio]{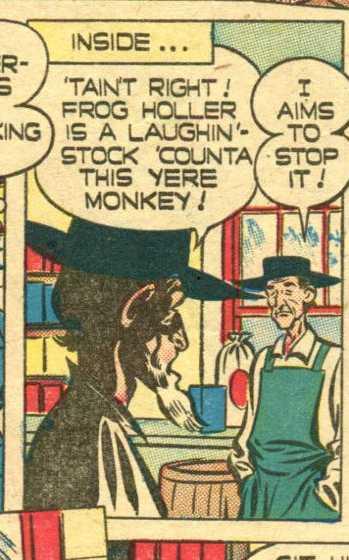}\par{\footnotesize\color{CaseMuted}Panel 1}
\end{minipage}\hfill
\begin{minipage}[t]{.485\linewidth}
\centering
\includegraphics[width=\linewidth,height=1.5in,keepaspectratio]{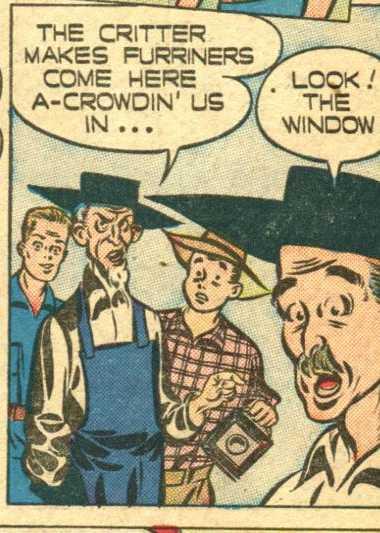}\par{\footnotesize\color{CaseMuted}Panel 2}
\end{minipage}\par\smallskip
\noindent\begin{minipage}[t]{.485\linewidth}
\centering
\includegraphics[width=\linewidth,height=1.5in,keepaspectratio]{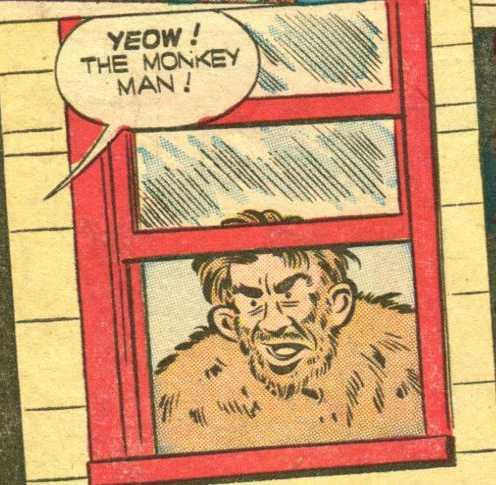}\par{\footnotesize\color{CaseMuted}Panel 3}
\end{minipage}\hfill
\begin{minipage}[t]{.485\linewidth}
\centering
\includegraphics[width=\linewidth,height=1.5in,keepaspectratio]{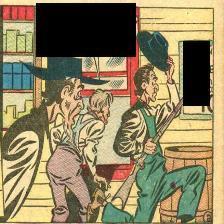}\par{\footnotesize\color{CaseMuted}Panel 4}
\end{minipage}\par
\casefield{Question}
How does the encounter at the window relate to the apron-wearing man's earlier objective and his alarmed reaction in the last panel?
\caseoption{A}{He wants to stop outsiders from crowding the town. He takes the figure at the window for another unwanted visitor, so his alarm responds to the influx he has been complaining about.}
\caseoption{B}{He wants to stop the monkey-related nuisance. The window brings him face-to-face with its apparent source, so his alarm interrupts the resolve he expressed before the encounter.}
\caseoption{C}{He wants to keep outsiders from seeing the monkey. Its appearance at the window exposes the attraction, so his alarm responds to the failure of his effort to conceal it.}
\noindent\begin{minipage}{\linewidth}
\caseoption{D}{He wants to stop the monkey-related nuisance. Its apparent source comes within reach at the window, so his alarm accompanies an opportunity that moves his intended intervention forward.}
\caseanswer{Qwen3.8-Flash}{D}{B}
\end{minipage}\par

\end{benchmarkcase}

\begin{benchmarkcase}[breakable=true]{CaseImage}{06}{Diagnostic Evidence}
\noindent\begin{minipage}{\linewidth}
\centering
\includegraphics[width=\linewidth,height=2.0in,keepaspectratio]{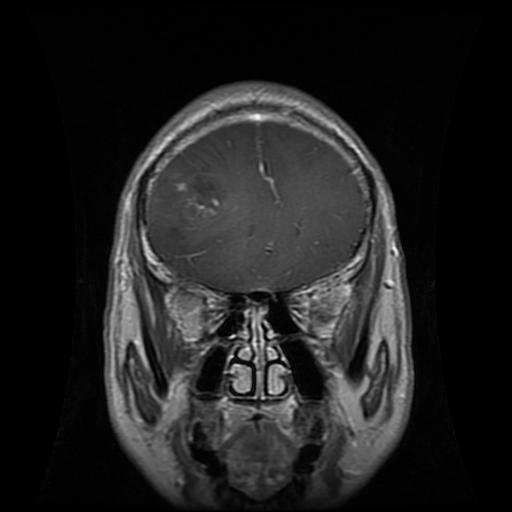}
\end{minipage}\par\smallskip
\casefield{Context}
This is a hypothetical untreated presentation in which infection and a necrotic neoplasm remain plausible, and tissue acquisition is already clinically indicated. The target is the tissue at pixel (181, 187) in the displayed 512 \(\times\) 512 image, measured from the upper-left corner, with x increasing rightward and y downward. Full MRI and procedural planning will establish image registration, the corresponding sampling territories, and safe access; this single image does not establish a safe trajectory or sequence-specific tissue identity. The proposed territories are the irregular peripheral tissue bordering the enclosed darker material and the enclosed material itself. All packages use the same procedure and timing, obtain architecture-preserving tissue where feasible, and provide appropriately handled portions for histopathology and microbiology. For separately processed specimens, 'inflammation with concordant microbiology' means inflammatory histopathology and corroborating microbiological evidence from that same registered territory, not an isolated potentially contaminating culture. All described results are possible future outcomes, not findings already obtained.

\casefield{Question}
Which prospective sampling-and-interpretation package best determines what is supported in the target tissue, while distinguishing that conclusion from coexistence of processes elsewhere in the lesion?
\caseoption{A}{Keep peripheral and enclosed specimens separate for histopathology and microbiology. If peripheral tissue contains viable tumor and the enclosed specimen has inflammation with concordant microbiology, this supports infection in the target. With the reverse regional findings, it supports tumor in the target. Either outcome supports lesion-level coexistence, not necessarily both processes within the target; an additional unsampled process remains possible.}
\caseoption{B}{Pool peripheral and enclosed specimens for histopathology and microbiology, retaining their image-registered collection records. If the pooled material contains viable tumor and inflammation with concordant microbiology, this supports both processes within the target. If it contains only inflammation with concordant microbiology, this supports infection in the sampled target. An additional unsampled process remains possible in either outcome.}
\caseoption{C}{Keep peripheral and enclosed specimens separate for histopathology and microbiology. If peripheral tissue contains viable tumor and the enclosed specimen has inflammation with concordant microbiology, this supports tumor in the target. With the reverse regional findings, it supports infection in the target. Either outcome supports lesion-level coexistence, not necessarily both processes within the target; an additional unsampled process remains possible.}
\noindent\begin{minipage}{\linewidth}
\caseoption{D}{Keep peripheral and enclosed specimens separate for histopathology but pool their microbiology portions. If peripheral tissue contains viable tumor and inflammation and the pooled microbiology identifies a compatible organism, this supports both processes within the target. If peripheral tissue contains inflammation without demonstrated tumor and pooled microbiology is concordant, this supports infection in the sampled target. An additional unsampled process remains possible in either outcome.}
\caseanswer{Qwen3.8-Flash}{A}{C}
\end{minipage}\par

\end{benchmarkcase}

\begin{benchmarkcase}{CaseImage}{07}{Animal Reasoning}
\noindent\begin{minipage}{\linewidth}
\centering
\includegraphics[width=\linewidth,height=1.7in,keepaspectratio]{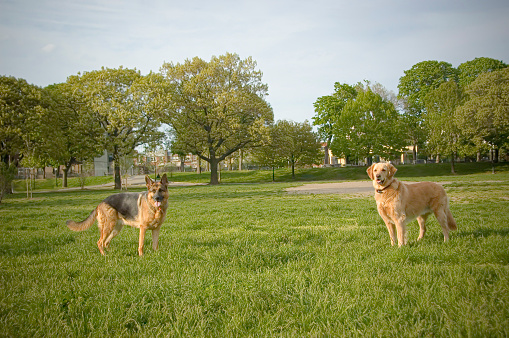}
\end{minipage}\par\smallskip
\casefield{Question}
A sheep-management demonstration will feature the historical work of the breed represented by the dog on the left. In one setting, sheep scattered across an enclosed pasture must be collected and moved through a gate. In the other, a cohesive flock must keep feeding on an unfenced grazing strip beside cultivated land. Which assignment and method most specifically illustrate this breed's characteristic historical contribution?
\caseoption{A}{Assign the enclosed pasture: gather the sheep into a group, then work behind them to move them through the gate, demonstrating a tradition of driving livestock.}
\caseoption{B}{Assign the grazing strip: work ahead of the flock and guide it along the strip to successive feeding patches, demonstrating a tradition of leading livestock.}
\caseoption{C}{Assign the enclosed pasture: sweep behind the scattered sheep and gather them toward the handler for passage through the gate, demonstrating a tradition of fetching livestock.}
\noindent\begin{minipage}{\linewidth}
\caseoption{D}{Assign the grazing strip: patrol its edges and turn sheep back when they stray toward the crops while letting them feed, demonstrating a tradition of tending livestock.}
\caseanswer{Qwen3.8-Flash}{C}{D}
\end{minipage}\par

\end{benchmarkcase}

\casegroup{Audio--Text Cases}{For each of the five Audio--Text benchmarks, we present one example that Qwen3.8-Omni-Flash answered incorrectly. Each example includes an audio transcript or excerpt, the question and answer options, and textual context where applicable. The bottom row shows the model prediction in red and the reference answer in green. Transcripts are provided for readability; the model receives audio inputs when answering.}

\begin{benchmarkcase}[breakable=true]{CaseAudio}{08}{Lyric Interference}
\casefield{Audio transcript}
\textbf{Organizer.} Before we finish family dance practice: everyone who partnered the child in this round must attend the repeat, except anyone who has already completed the separate demonstration.\par
\textbf{Judy.} I offered to do that demonstration, but I haven't actually done it. Jack, did you finish yours?\par
\textbf{Jack.} Yes, I completed mine before practice started. I may not know those flash dances, but I'm no slouch on the dance floor.\par
\textbf{Organizer.} Good. My partner report lists exactly two people for your child: Judy and our guest, Peter. Nobody else.\par
\textbf{Child.} Mom, I'm hungry. Can we get something after this?\par
\textbf{Judy.} Soon. First, listen while we check the practice report.\par
\textbf{Child.} There's a mistake in it. My second partner wasn't Peter; it was my dad, Jack. Replace Peter with Jack. Mom was my first partner, so Judy is right.\par
\textbf{Organizer.} Thanks for correcting the report. I'll check it against the attendance rule before we start the repeat.\par

\casefield{Question}
Under the organizer's rule, who is required to attend the repeat dance?
\caseoption{A}{Judy only.}
\caseoption{B}{Judy and Peter.}
\caseoption{C}{Jack only.}
\noindent\begin{minipage}{\linewidth}
\caseoption{D}{Judy and Jack.}
\caseanswer{Qwen3.8-Omni-Flash}{B}{A}
\end{minipage}\par

\end{benchmarkcase}

\begin{benchmarkcase}[breakable=true]{CaseAudio}{09}{Audio Misdirection}
\casefield{Audio transcript}
\textbf{Narrator.} Consider a hypothetical production exercise inspired by The World According to Dick Cheney, a documentary combining interviews and narration. This incident is invented, not a report about the actual filmmakers.

A preview team received confidentiality training. The rule was that no interview excerpt could be released until the producer authorized that particular excerpt. Once authorized, that excerpt could be posted publicly.

An audit of the restricted period found that the team had withheld every excerpt and that no unauthorized copies had escaped.

Meanwhile, editors were balancing interview passages with narration and preparing publicity before the scheduled debut. One interview passage was also intended to surprise viewers at the full screening.

Later, the producer approved a promotional excerpt containing that passage and instructed the preview team to post it. The team posted only the approved material. Viewers then reshared the public excerpt, so the passage reached people before the full screening and the planned surprise was lost.

Further promotional postings can be paused while the production group reviews its release policy and decides which passages to approve next.\par

\casefield{Context}
Production commentary: The audit is reassuring: the preview team initially withheld everything, and the producer did authorize the eventual post. Viewers, rather than the team, then spread it further. Pausing additional excerpts is therefore sensible. Still, confidentiality training was meant to keep interview material from spoiling the screening, and that same problem has now returned. Permission explains why the team posted the excerpt; it does not explain away the loss of confidentiality. The initial success evidently did not last through the promotional phase. Before producer-approved posting resumes, the team should repeat the training so that its earlier protection is restored.
\casefield{Question}
Which proposed next step and justification are best supported by the recording?
\caseoption{A}{Pause further excerpts and review which material is approved for promotion, because the team observed the restriction but the authorized posting enabled the surprise to circulate.}
\caseoption{B}{Pause further excerpts and investigate a leak during the restricted period, because the later resharing shows that the audit missed an unauthorized release.}
\caseoption{C}{Resume approved promotion without changing excerpt selection, because the team observed the restriction and producer authorization removed the relevant risk to the surprise.}
\noindent\begin{minipage}{\linewidth}
\caseoption{D}{Pause further excerpts and repeat the confidentiality training before approved promotion resumes, because the lost surprise shows that the earlier training did not hold.}
\caseanswer{Qwen3.8-Omni-Flash}{D}{A}
\end{minipage}\par

\end{benchmarkcase}

\begin{benchmarkcase}[breakable=true]{CaseAudio}{10}{Target-Speaker Tracking}
\textit{Listening order: reference voice, then two conversation clips.}\par
\casefield{Audio transcript}
\textbf{Reference voice (Person B).} I left my umbrella by the door. Could you bring it over when you come back?\par
\textbf{Person A.} Did you read the letter in the paper about smoking? It focused on cinemas and theaters. The writer must be crazy.\par
\textbf{Person B.} I don't think so. I believe the government should do everything possible to discourage smoking.\par
\textbf{Person A.} Maybe, but a ban on cigarettes won't work. I'd restrict smoking only in restaurants and pubs, not in those entertainment venues. People spend longer over a meal or a drink, and that's where I'd draw the line.\par
\textbf{Person B.} No, you can't make cigarettes illegal overnight. But the government could prohibit smoking in those two venues.\par
\textbf{Person C.} I'd ban cigarette sales instead, while letting cinemas and theaters decide their own smoking rules. Shopkeepers would have to stop selling cigarettes, but the venue owners could decide what happens inside. That's the approach I'd support, rather than more rules about particular places.\par

\par\noindent\begin{minipage}{\linewidth}
\casefield{Question}
The first clip is a voice reference. The remaining clips form a conversation. Which policy does the speaker matching the reference voice support?
\caseoption{A}{Keep cigarettes legal, but ask cinemas and theaters to discourage smoking without prohibiting it.}
\caseoption{B}{Keep cigarettes legal, but prohibit smoking in public venues such as cinemas and theaters.}
\caseoption{C}{Keep cigarettes legal, but restrict smoking only in restaurants and pubs, not cinemas or theaters.}
\noindent\begin{minipage}{\linewidth}
\caseoption{D}{Ban cigarette sales, but let cinemas and theaters decide whether to allow smoking.}
\caseanswer{Qwen3.8-Omni-Flash}{D}{B}
\end{minipage}\par
\end{minipage}
\end{benchmarkcase}

\begin{benchmarkcase}[breakable=true]{CaseAudio}{11}{Audio Answerability}
\casefield{Audio transcript (excerpt)}
\textit{[\ldots]}\par
\textbf{Mon.} Sure. Ross, can I give Mom and Dad's toast this year instead of you?\par
\textbf{Ross.} Absolutely. I'll introduce you.\par
\textbf{Phoebe.} That was a lovely party tonight, Mon. Your dinner toast really moved your parents.\par
\textbf{Mon.} Thanks. My brother gave me a wonderful introduction.\par
\textbf{Ross.} Just an introduction, though. I handed you the microphone without proposing a toast myself.\par
\textbf{Phoebe.} I was with you both from the start through cake. Yours was the only toast in that stretch. Then I left to collect Parker from the station.\par
\textbf{Joey.} That's when I started clearing the dining room. I stayed there until your parents left. There weren't any more toasts in that room.\par
\textbf{Mon.} I stayed inside helping Joey. Ross took Mom and Dad onto the terrace while you were gone, Phoebe.\par
\textbf{Phoebe.} When we got back, your parents were leaving. Parker caught them at the door to say goodbye.\par

\casefield{Question}
Does the conversation establish that Mon was the only one of the two siblings to give a toast during the entire anniversary celebration?
\caseoption{A}{Yes, it establishes that only Mon gave a toast.}
\caseoption{B}{It does not establish whether only Mon gave a toast.}
\noindent\begin{minipage}{\linewidth}
\caseoption{C}{No, it establishes that both siblings gave a toast.}
\caseanswer{Qwen3.8-Omni-Flash}{A}{B}
\end{minipage}\par

\end{benchmarkcase}

\begin{benchmarkcase}[breakable=true]{CaseAudio}{12}{Speaker Attribution}
\textit{Listening order: reference voice, then one conversation clip.}\par
\casefield{Audio transcript (excerpt)}
\textbf{Reference voice (Morgan).} Let me take a moment to think about that.\par
\textbf{Morgan.} I told everyone in class Ross slept with dinosaurs. Obviously a joke! Now his rumor about me has people calling to congratulate me.\par
\textbf{Casey.} I already called Ross privately about that rumor. He laughed, but I asked him to stop spreading it.\par
\textbf{Alex.} I've already posted a public correction. I'm helping with the calls, and people keep asking whether his claim is true.\par
\textbf{Morgan.} I've drafted a correction email to the whole class. It's still unsent; I'm not sure sending it would help.\par
\textit{[\ldots]}\par
\textbf{Alex.} I'm considering calling him privately too. Maybe talking directly would finally persuade him to clear things up.\par
\textbf{Morgan.} Actually, I'll leave that draft unsent. I'll call Ross privately, but only if the congratulatory calls continue tomorrow.\par
\textbf{Casey.} I'll post my correction only if the class organizer approves it. Otherwise, no post.\par
\textbf{Alex.} Forget my private call. Instead, I'll email the class only if more congratulatory calls arrive tomorrow.\par

\casefield{Question}
Listen to the reference voice, then the conversation. Which pair correctly combines something that speaker has already done with the response they still intend to make under a stated condition?
\caseoption{A}{Posted a public correction; will email the class if more congratulatory calls arrive tomorrow.}
\caseoption{B}{Sent a class-wide correction email; will call Ross privately if congratulatory calls continue tomorrow.}
\caseoption{C}{Called Ross privately; will post a public correction if the class organizer approves.}
\noindent\begin{minipage}{\linewidth}
\caseoption{D}{Made the dinosaur joke to classmates; will call Ross privately if congratulatory calls continue tomorrow.}
\caseanswer{Qwen3.8-Omni-Flash}{A}{D}
\end{minipage}\par

\end{benchmarkcase}

\casegroup{Image--Audio--Text Cases}{For each of the four Image--Audio--Text benchmarks, we present one example that Qwen3.8-Omni-Flash answered incorrectly. Each example includes the original images, an audio transcript or excerpt, and the written question and answer options. Multiple images are displayed in their original input order. The bottom row compares the model prediction in red with the reference answer in green.}

\begin{benchmarkcase}{CaseOmni}{13}{Scene Misdirection}
\noindent\begin{minipage}{\linewidth}
\centering
\includegraphics[width=\linewidth,height=1.7in,keepaspectratio]{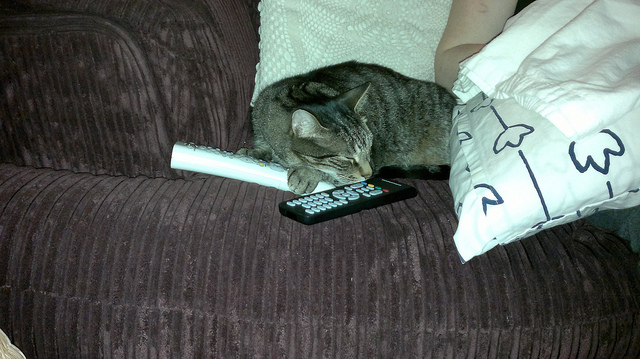}
\end{minipage}\par\smallskip
\casefield{Audio transcript}
\textbf{Speaker 1.} That cat has settled into such a snug little spot with the two remotes. They're all clustered right beside the heart pillow, tucked behind it together. The pillow partly covers the whole little group.\par
\textbf{Speaker 2.} Yes, they're so close together there. The cat and both remotes are partly hidden by that same pillow, like one cozy bundle on the couch.\par

\casefield{Question}
The resting cat and two remotes form a close cluster beside the heart-patterned pillow, tucked together behind it so that the pillow partly covers all three.

Which members of this group are partly hidden from view by the foreground heart-patterned pillow?
\caseoption{A}{The cat and both remotes.}
\caseoption{B}{The cat and the black remote only.}
\caseoption{C}{The cat alone.}
\noindent\begin{minipage}{\linewidth}
\caseoption{D}{The cat and the white remote only.}
\caseanswer{Qwen3.8-Omni-Flash}{A}{C}
\end{minipage}\par

\end{benchmarkcase}

\begin{benchmarkcase}{CaseOmni}{14}{Art-Historical Reasoning}
\noindent\begin{minipage}{\linewidth}
\centering
\includegraphics[width=\linewidth,height=2.3in,keepaspectratio]{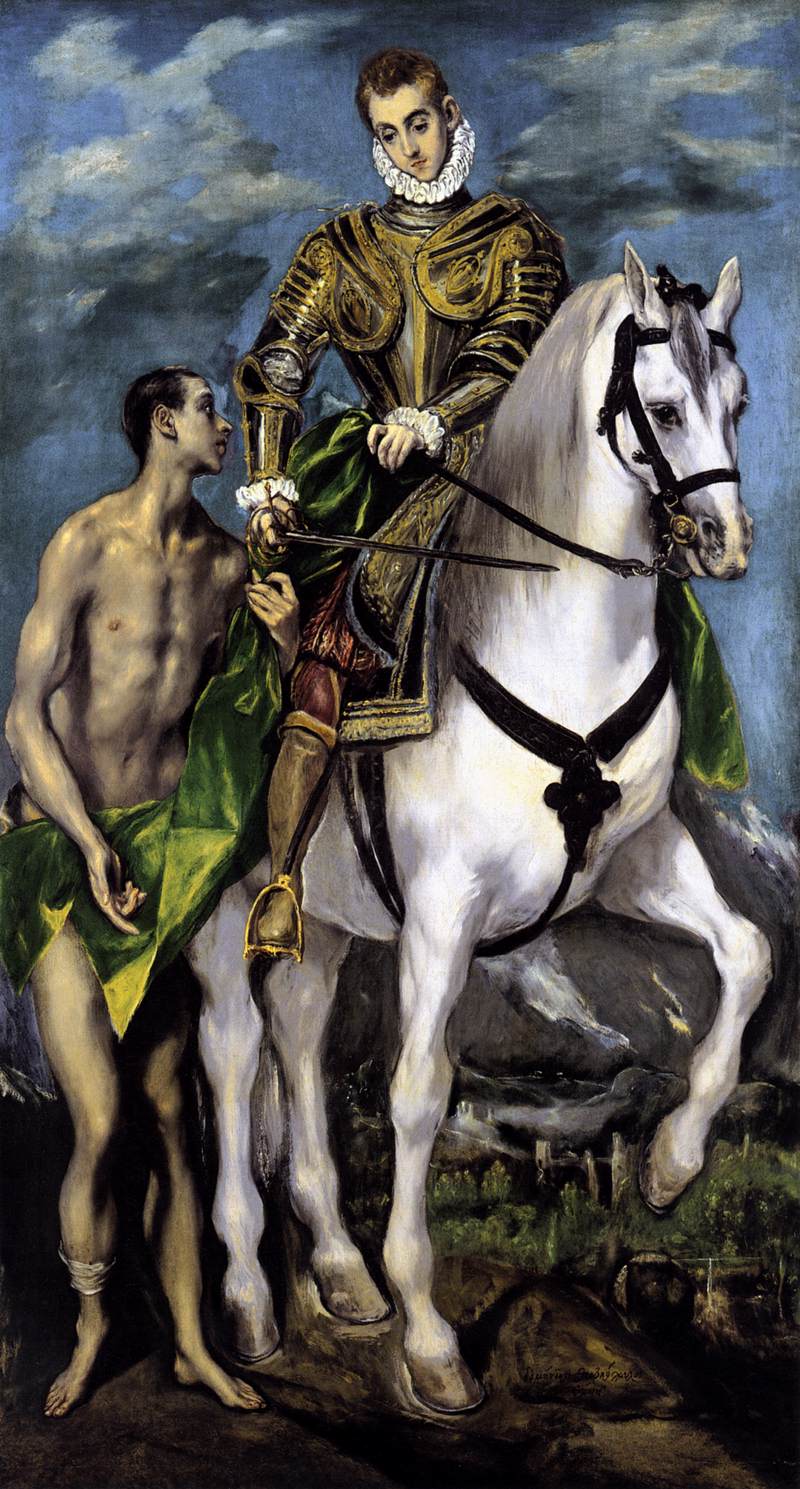}
\end{minipage}\par\smallskip
\casefield{Audio transcript}
\textbf{Companion.} Elena, that green really stands out against the cloudy sky.\par
\textbf{Elena.} It does. Which painting faced this one across the chapel in its original installation?\par

\casefield{Question}
Answer Elena's question.
\caseoption{A}{Saint Joseph and the Christ Child}
\caseoption{B}{Coronation of the Virgin}
\noindent\begin{minipage}{\linewidth}
\caseoption{C}{Madonna and Child with Saint Martina and Saint Agnes}
\caseanswer{Qwen3.8-Omni-Flash}{A}{C}
\end{minipage}\par

\end{benchmarkcase}

\begin{benchmarkcase}[breakable=true]{CaseOmni}{15}{Visual Coverage}
\noindent\begin{minipage}{\linewidth}
\centering
\includegraphics[width=\linewidth,height=1.7in,keepaspectratio]{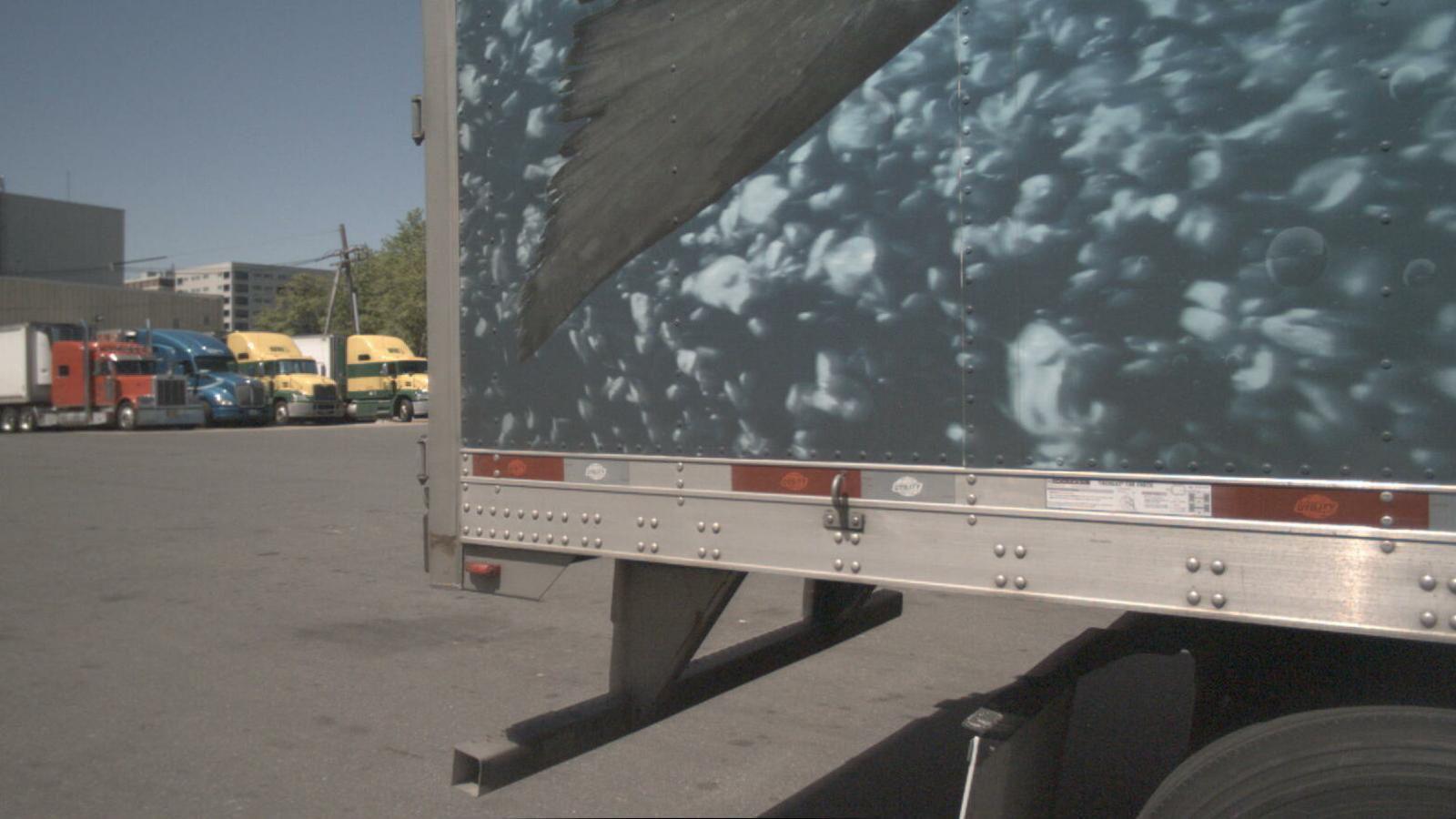}\par{\footnotesize\color{CaseMuted}Image 1}
\end{minipage}\par\smallskip
\noindent\begin{minipage}{\linewidth}
\centering
\includegraphics[width=\linewidth,height=1.7in,keepaspectratio]{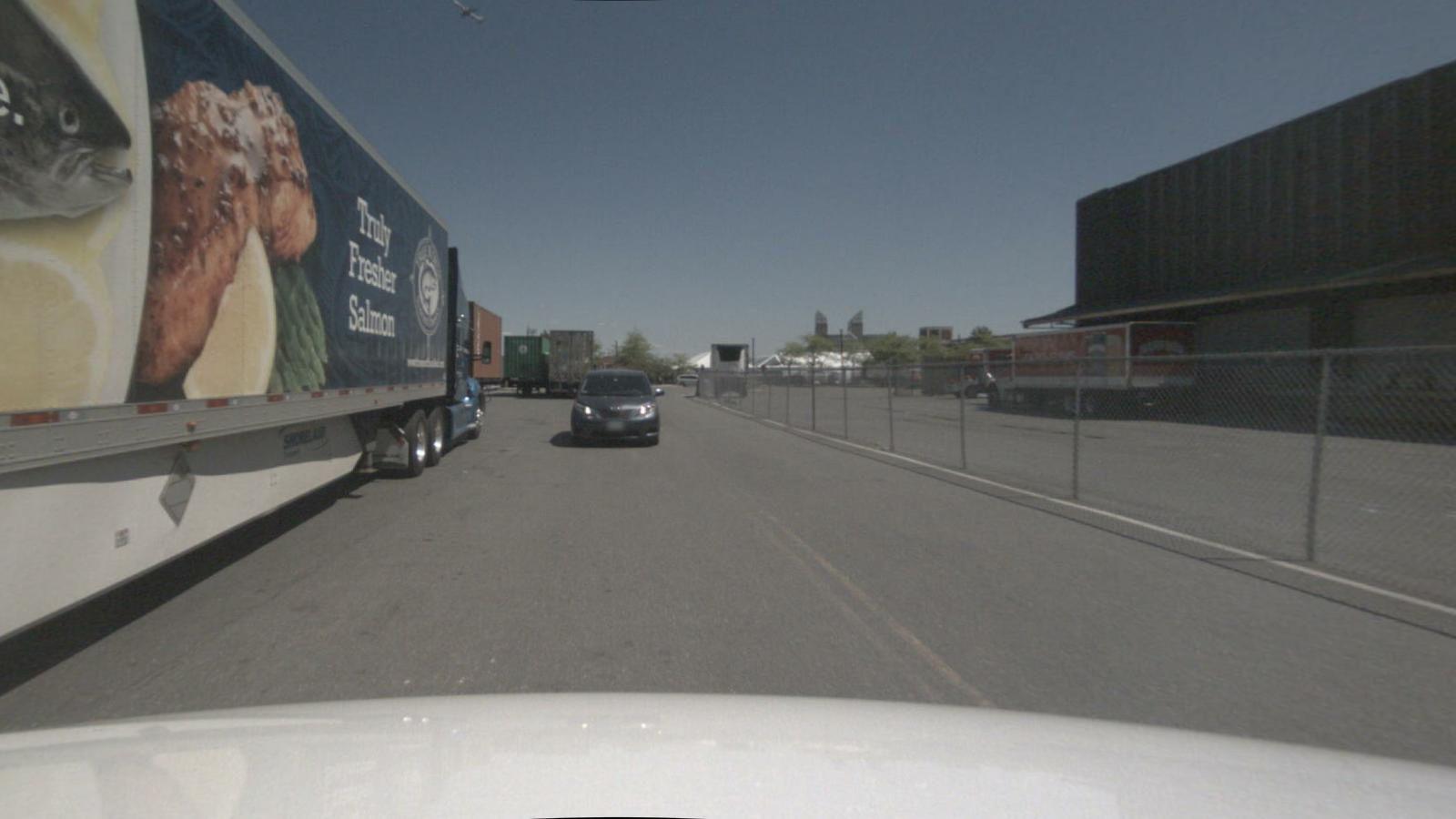}\par{\footnotesize\color{CaseMuted}Image 2}
\end{minipage}\par\smallskip
\noindent\begin{minipage}{\linewidth}
\centering
\includegraphics[width=\linewidth,height=1.7in,keepaspectratio]{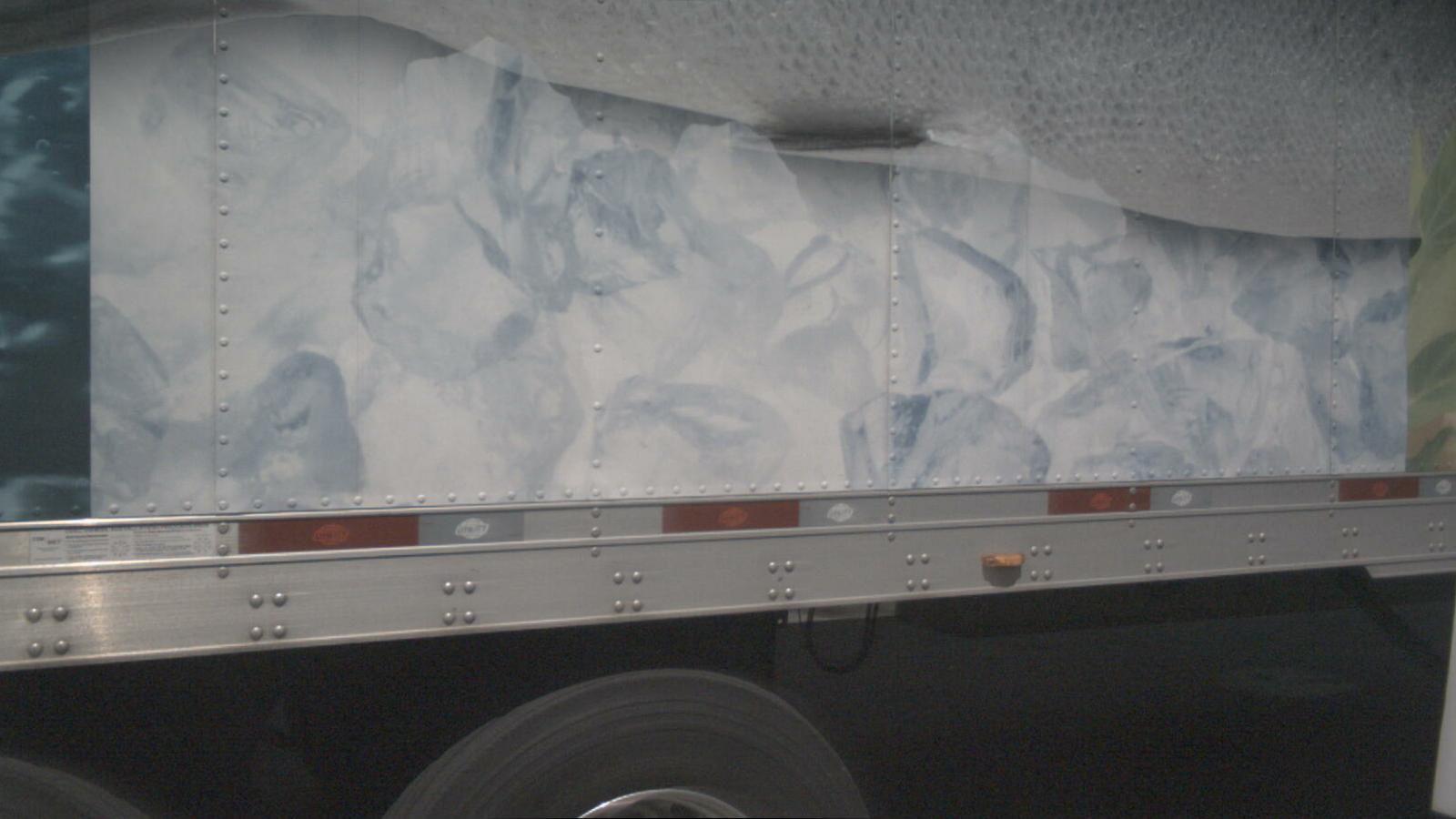}\par{\footnotesize\color{CaseMuted}Image 3}
\end{minipage}\par\smallskip
\casefield{Audio transcript}
\textbf{Companion.} Maya, those seafood graphics take up almost the whole side of the trailer.\par
\textbf{Maya.} They do. Let's look through the three photos together.\par
\textbf{Maya.} For the trailer and tires defined in the scope note, do these photographs confirm that at least one wheel chock is positioned against a tire, confirm that none is, or leave it unresolved?\par

\casefield{Question}
Answer Maya's question in the audio using the three photographs together.

Scope note: Inspect only the nearby trailer with the large seafood graphics. Determine whether at least one wheel chock is positioned against any of that trailer's road tires, on either side. Exclude the tractor and all background vehicles. A wheel chock is a block placed against a tire to prevent rolling.
\caseoption{A}{Confirmed absent}
\caseoption{B}{Confirmed present}
\noindent\begin{minipage}{\linewidth}
\caseoption{C}{Unresolved}
\caseanswer{Qwen3.8-Omni-Flash}{A}{C}
\end{minipage}\par

\end{benchmarkcase}

\begin{benchmarkcase}[breakable=true]{CaseOmni}{16}{Visual Detail QA}
\noindent\begin{minipage}{\linewidth}
\centering
\includegraphics[width=\linewidth,height=1.7in,keepaspectratio]{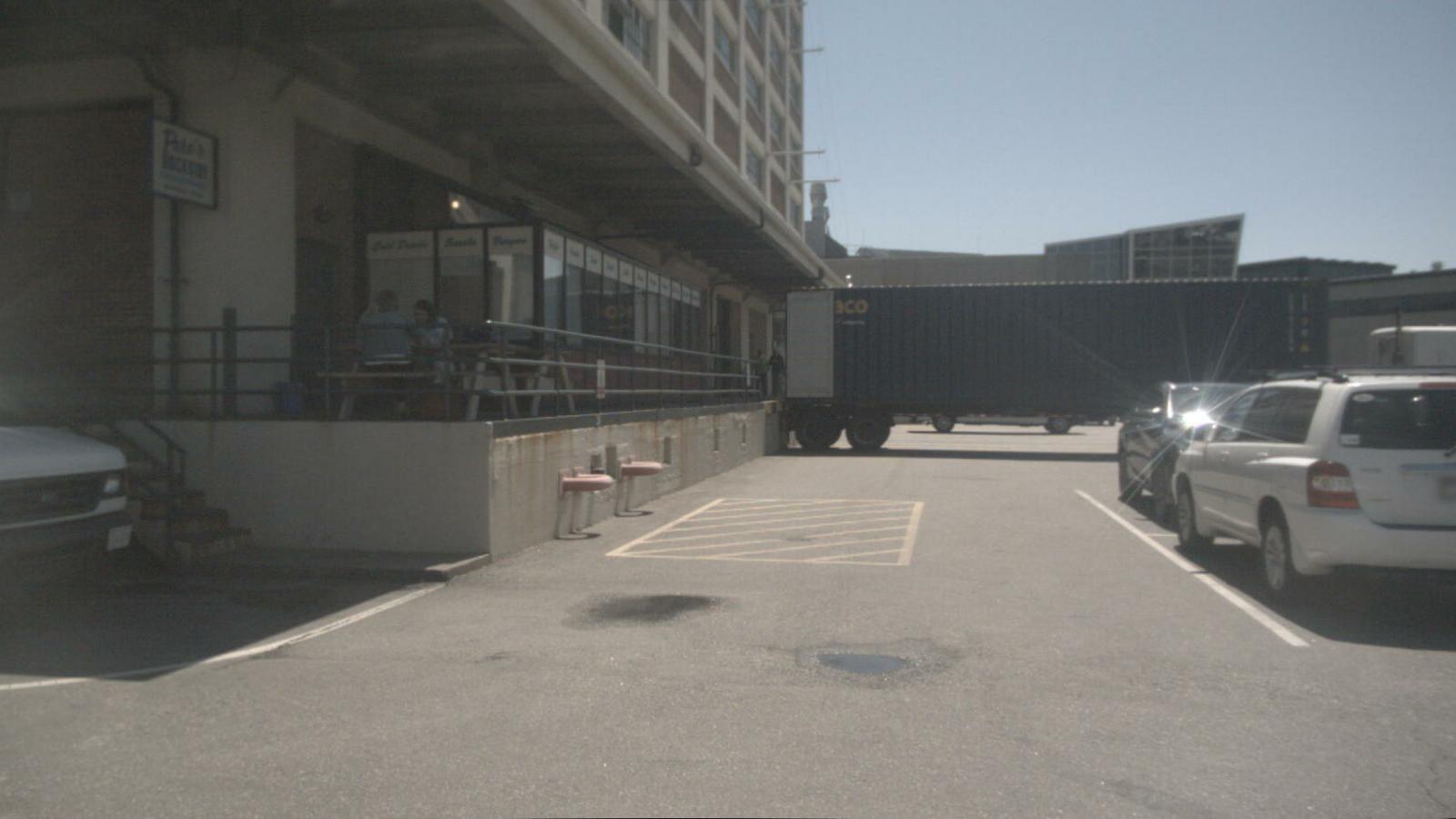}\par{\footnotesize\color{CaseMuted}Image 1}
\end{minipage}\par\smallskip
\noindent\begin{minipage}{\linewidth}
\centering
\includegraphics[width=\linewidth,height=1.7in,keepaspectratio]{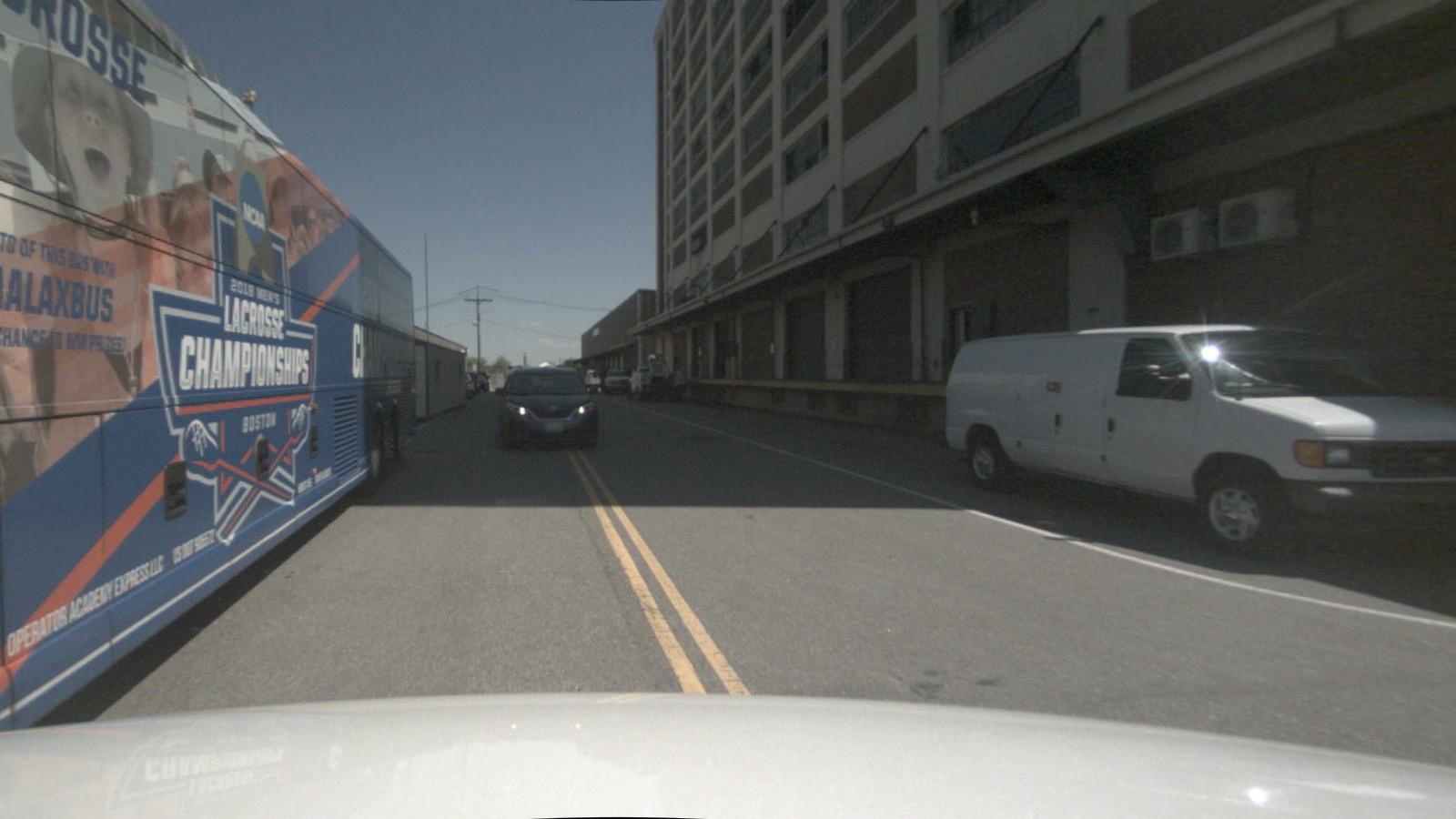}\par{\footnotesize\color{CaseMuted}Image 2}
\end{minipage}\par\smallskip
\casefield{Audio transcript}
\textbf{Casey.} Jordan, these industrial buildings have quite a presence from street level.\par
\textbf{Jordan.} They do. Looking at both photographs, consider the white van whose front is cut off by the left edge of photograph one, beside the short staircase. Does its photographed cargo side behind the cab have no windows, exactly one window, or two or more windows? Count only windows in that side behind the cab, excluding the windshield, cab-door glass, and rear-facing doors.\par

\casefield{Question}
Answer Jordan's question using the two photographs.
\caseoption{A}{The photographed cargo side has no windows.}
\caseoption{B}{Cannot determine from the supplied photographs.}
\caseoption{C}{The photographed cargo side has two or more windows.}
\noindent\begin{minipage}{\linewidth}
\caseoption{D}{The photographed cargo side has exactly one window.}
\caseanswer{Qwen3.8-Omni-Flash}{B}{A}
\end{minipage}\par

\end{benchmarkcase}

\clearpage
\endgroup

\end{document}